\documentclass[sn-mathphys,Numbered]{sn-jnl-arxiv}

\usepackage{natbib}
\usepackage{graphicx}%
\usepackage{multirow}%
\usepackage{amsmath,amssymb,amsfonts}%
\usepackage{amsthm}%
\usepackage{mathrsfs}%
\usepackage{xcolor}%
\usepackage{textcomp}%
\usepackage{manyfoot}%
\usepackage{booktabs}%
\usepackage{algorithm}%
\usepackage{algorithmicx}%
\usepackage{algpseudocode}%
\usepackage{listings}%

\usepackage{fancyvrb,fvextra}

\newcommand{\blue}[1]{{\color{blue}#1}}

\usepackage[ampersand]{easylist}
\ListProperties(Hide2=1,Hide3=2,Progressive*=.5cm,Numbers3=l, Numbers4=r,FinalMark3={)})
\usepackage{etoolbox}
\AtBeginEnvironment{easylist}{\ListProperties(Start1=1)}

\usepackage[dvipsnames]{xcolor}

\definecolor{cblue}{RGB}{8, 85, 153}
\definecolor{darkgreen}{rgb}{0.0, 0.5, 0.0}

\usepackage[capitalise]{cleveref}
\usepackage{float}
\usepackage{placeins} 
\usepackage{adjustbox}

\usepackage{makecell}
\usepackage[ampersand]{easylist}
\usepackage{todonotes}
\usepackage{caption}
\usepackage{xspace}
\usepackage{subcaption}

\newcommand{\method}{HACHI\xspace}

\usepackage{adjustbox}
\usepackage{textcomp}

\usepackage{titlesec}
\titleformat{\section}
  {\normalfont\large\bfseries}
  {\thesection}{1em}{}
\makeatletter
\renewcommand\paragraph{\@startsection{paragraph}{4}{\z@}%
  {1ex \@plus 1ex \@minus .2ex}%
  {-1em}%
  {\normalfont\normalsize\bfseries}}
\makeatother

\usepackage{setspace}
\begin{document}

\title[HACHI]{Human-AI Co-design for Clinical Prediction Models}

\author[1,*,†]{\fnm{Jean} \sur{Feng}}\email{}
\author[1,*]{\fnm{Avni} \sur{Kothari}}\email{}
\author[1]{\fnm{Patrick} \sur{Vossler}}\email{}
\author[1]{\fnm{Andrew} \sur{Bishara}}\email{}
\author[1]{\fnm{Lucas} \sur{Zier}}\email{}
\author[1]{\fnm{Newton} \sur{Addo}}\email{}
\author[1]{\fnm{Aaron} 
\sur{Kornblith}}\email{}
\author[2]{\fnm{Yan Shuo} \sur{Tan}}\email{}
\author[3]{\fnm{Chandan} \sur{Singh}}\email{}

\affil[1]{\orgdiv{University of California, 
San Francisco}}
\affil[2]{\orgdiv{National University of Singapore}}
\affil[3]{\orgdiv{Microsoft Research}\vspace{4pt}}
\affil[*]{Equal contribution}
\affil[†]{Corresponding author: jean.feng@ucsf.edu}

\abstract{
Developing safe, effective, and practically useful clinical prediction models (CPMs) traditionally requires iterative collaboration between clinical experts, data scientists, and informaticists.
This process refines the often small but critical details of the model building process, such as which features/patients to include
and how clinical categories should be defined.
However, this traditional collaboration process is extremely time- and resource-intensive,
resulting in only a small fraction of CPMs reaching clinical practice.
This challenge intensifies when teams attempt to reliably incorporate information from unstructured clinical notes, which can contain an essentially infinite number of concepts.
To address this challenge, we introduce HACHI, an iterative human-in-the-loop framework that uses AI agents to accelerate the development of fully interpretable CPMs by enabling the exploration of concepts in clinical notes.
HACHI alternates between (i) an AI agent rapidly exploring and evaluating candidate concepts in clinical notes and (ii) clinical and domain experts providing feedback to improve the CPM learning process.
HACHI defines concepts as simple yes-no questions that are used in linear models, allowing the clinical AI team to transparently review, refine, and validate the CPM learned in each round.
In two real-world prediction tasks (acute kidney injury and traumatic brain injury), HACHI outperforms existing approaches, surfaces new clinically relevant concepts not included in commonly-used CPMs, and improves model generalizability across clinical sites and time periods.
Furthermore, HACHI reveals the critical role of the clinical AI team, such as directing the AI agent to explore entire categories of concepts that it had not previously considered, adjusting the granularity of concepts it considers, changing the objective function to better align with the clinical objectives, and identifying issues of data bias and leakage.
Code for HACHI is available at \url{http://github.com/jjfenglab/HACHI}.
}
\keywords{Large language models, Electronic health records, Concept Bottleneck, Human-AI Interaction}

\maketitle

\begin{figure}
    \centering
    \makebox[\textwidth]{
    \includegraphics[width=1.15\textwidth]{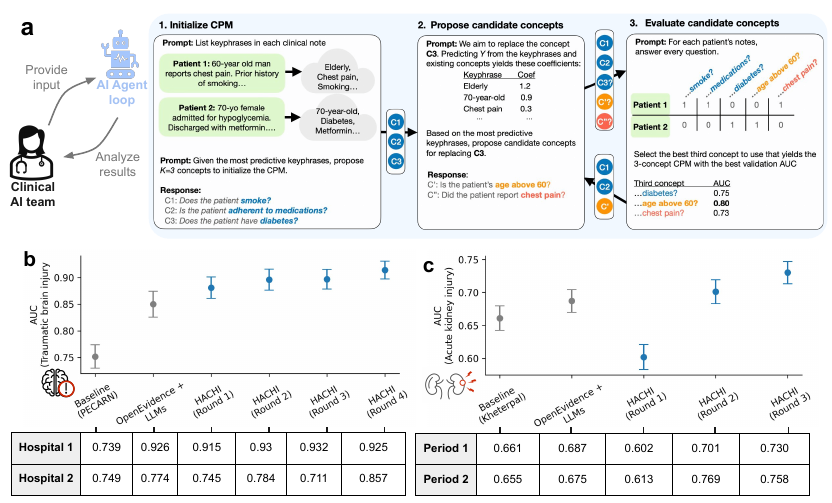}}
    \caption{\textbf{The \method framework uses LLMs with humans-in-the-loop to build an effective clinical prediction model (CPM).}
    (a) HACHI is composed of an outer loop where the clinical AI team provides guidance and feedback to the AI agent on how to learn a CPM, and an inner loop where the AI agent follows instructions from the clinical AI team to find $k$ concepts (formally defined as yes/no questions) that maximize the CPM's predictive accuracy.
    The inner AI-guided CPM learning procedure is broken down into three steps: 1. Initialize the CPM by brainstorming clinical concepts from keyphrases extracted out of clinical notes; 2. Propose candidate concepts by analyzing which keyphrases are most associated with the outcome of interest; 3. Evaluate the candidate concepts by annotating each concept and selecting the best-performing concept(s). Steps 2 and 3 are repeated until convergence.
    Each round, the clinical AI team analyzes the results from the AI-guided CPM learning procedure and provides feedback on how the procedure can be improved, such as by modifying the prompts and clarifying which concepts are and are not of interest.
    \method improves over baselines and across rounds for two real-world clinical prediction tasks:
    (b) diagnosis of traumatic brain injury (TBI), where performance is evaluated in terms of AUC with respect to the overall population (shown in plot) and  stratified across two sites (shown in table).
    (c) development of acute kidney injury (AKI), where performance is evaluated in an internal validation set (Period 1, shown in plot and table) and a later, temporally disjoint test dataset (Period 2, shown in table).
    Error bars show standard errors.
    }
    \label{fig:intro}
\end{figure}


\section{Introduction}
\label{sec:intro}

Clinical prediction models (CPMs) translate routinely collected clinical information into structured predictions, enabling clinicians to apply shared expertise in a more consistent manner across patients.
While there is growing interest in highly complex black-box CPMs, simple fully-interpretable CPMs, such as rule-based scores like SOFA for ICU mortality \citep{Ranzani2025-yh} or PECARN and NEXUS for traumatic brain injury \citep{kuppermann2009identification}, remain the most popular in clinical practice due to a number of major advantages.
First and foremost, these models can be easily implemented by clinicians at the bedside using simple mental arithmetic or a quick chart evaluation, which facilitates their integration into the clinical workflow \citep{Teasdale1974-wx, Laupacis1997-cj}.
Furthermore, fully-interpretable models can be thoroughly audited by domain experts and, when necessary, modified to improve reliability, which promotes user trust \citep{Rudin2019-yb, Koh2020-kl}.
Consequently, when simple, fully-interpretable CPMs both meet clinical standards and achieve high performance, these tools are widely used by the clinical community \citep{Wells2000-sc, Chung2016-lg, McLendon2025-sa, Jain2025-tf}.

Despite their utility, very few such CPMs have achieved the level of performance and clinical sophistication needed for widespread clinical adoption~\citep{Damen2016-gw, Wynants2020-hl, Feng2023-ls}.
The main challenge is that ``the devil is in the details''---a CPM is determined by the \textit{numerous} details and decisions made in the CPM learning process, including which learning procedure was used, how the data was generated and compiled, which features are included, and how the model was evaluated \citep{van-Os2022-vv, Ramaswamy2023-qn, Obra2025-br}.
All of these factors must be carefully chosen, which generally requires
a lengthy and laborious collaboration
between clinicians and data scientists to iterate on the model learning procedure so that the final CPM both satisfies complex clinical requirements and meets desired levels of performance \citep{Mattei2020-ea, Ranzani2025-yh}.
To make matters worse, maximizing model performance often requires extending beyond tabular data in the Electronic Health Record (EHR) and analyzing unstructured clinical notes, as the latter often contains the most relevant and predictive features \citep{Jiang2023-sp,Seinen2022-zv, Seinen2025-dc}.
However, clinical notes cover an essentially \textit{infinite} number of concepts; for example, the concept of \textit{smoking} can be characterized by whether a patient currently smokes, has a history of smoking, is trying to quit smoking, and many more.
As such, it is infeasible for clinical AI teams to manually explore all the possible concepts for inclusion in the CPM learning process.

Recent developments in artificial intelligence (AI) suggest that AI agents can efficiently navigate and optimize many of these critical design decisions for learning fully-interpretable CPMs.
Indeed, recent research has shown that because large language models (LLMs) can now mimic complex clinical reasoning and accurately extract concepts from clinical notes \citep{Agrawal2022-me, Yang2022-ww, Guevara2024-ul}, AI agents can now analyze clinical notes, brainstorm useful features to include in a CPM, convert these features into simple tabular elements, and apply statistical tools to fit models~\citep{Oikarinen2023-ee, McInerney2023-ab, benara2025crafting, kim2024constructing, feng2024bayesian}.
To find concepts in clinical notes that are most predictive of the target of interest, these methods have an AI agent iteratively explore the infinite space of concepts based on its clinical knowledge, which can reveal novel predictive signals that are difficult to identify upfront \citep{feng2024bayesian,ludan2023interpretable}.
However, existing works only study the AI agent working in isolation.
Without external guidance from domain experts, the AI agent may learn concepts that are overly simplistic or lack clinical credibility, leading to CPMs that are promising but lack the clinical sophistication necessary for translation into clinical practice.

This work studies how clinical AI teams can collaborate with AI agents to improve the CPM learning process.
This cannot be achieved simply through one-time prompt tuning because of the so-called ``gulf of envisionment'' \citep{Subramonyam2024-at, Kothari2026-zo}: In the initial stage of building a CPM, the clinical AI teams often lack alignment and clarity on the exact task specifications, so it is often unclear how to best instruct an AI agent to achieve the team's goals.
To close this ``gulf of envisionment,'' the team must clarify to the AI agent what specific data to analyze, how exactly the outcome should be defined, what concepts to explore, what statistical tools to use, and more.
This can only be accomplished through an iterative, collaborative process, in which the human experts gradually understand the capabilities of the AI agent, gain clarity in how the task should be specified, and learn how to align the AI agent to learn the best possible CPM.
Through this process, team members can also build trust in the AI agent's learning process and, ultimately, the resulting CPM.

To this end, we introduce a \underline{H}uman+\underline{A}gent \underline{C}o-design framework for \underline{H}ealthcare \underline{I}nstruments~(\method)\footnote{HACHI references Hachikō (1923-1935), an Akita dog commemorated in Japan for his strong loyalty to his owner. The name highlights how the process of clinical AI teams learning to work with an AI agent is similar to the iterative process of owners learning to train their dogs.}.
In HACHI, the role of the AI agent is to convert unstructured clinical data into structured, reviewable concepts for learning the best-performing CPM, per instructions from the clinical AI team; the role of the clinical AI team is to analyze the results from each round using their high-level design and clinical reasoning to repeatedly refine the instructions to the AI agent.
As such, the AI agent does much of the ``heavy-lifting''---e.g., analyzing clinical notes, brainstorming candidate concepts, fitting statistical models, and selecting which concepts to include---while the clinical AI team provides overall, directional feedback on how the AI agent could improve its process (\cref{fig:intro}).
HACHI allows the clinical AI team to fully collaborate with the AI agent because every step of HACHI is interpretable.
We apply \method to two real-world clinical risk-prediction tasks---acute kidney injury and traumatic brain injury---to evaluate the efficacy of this human+AI agent co-design process.

\section{Results}

The goal of HACHI is to learn a CPM with $k$ concepts that are interpretable, clinically reasonable, and maximize predictive performance, given a dataset of clinical notes and corresponding labels.
Concepts are defined as answers to yes/no questions, such as ``Does a patient have...?''.
Because clinical AI teams often do not have a precise definition of this high-level goal at the beginning of a project, HACHI involves multiple rounds of fitting CPMs, each time with an AI-powered CPM learning procedure that is precisely specified but potentially imperfect.
The clinical AI team gathers new insights from each round, which are then used to improve how the AI-powered CPM learning procedure is run.
After multiple rounds, the learning procedure becomes better-aligned with the clinical AI team's goals, resulting in a CPM with the desired characteristics.

The AI-guided CPM learning procedure used in this work  (\cref{fig:intro} blue box) is a greedy hill-climbing procedure based on \citep{feng2024bayesian} that iteratively refines a $k$-concept CPM by cyclically traversing each concept position and replacing the incumbent concept when a better-performing concept is found. The AI agent iteratively brainstorms, evaluates, and selects CPM concepts per the following steps:
\begin{enumerate}
    \item \textbf{Initialize}: The AI agent creates a set of keyphrases to represent each observation, per instructions given in \texttt{KeyphrasePrompt}. The CPM is initialized with $k$ concepts per an \texttt{InitializationPrompt}, and the data is partitioned into a training set and a validation set.
    \item \textbf{Propose candidate concepts}: Using the training data, the AI agent fits a statistical model to identify the top keyphrases associated with the outcome. Combining this list with its prior knowledge, the AI agent proposes candidate concepts in the form of yes/no questions for inclusion in the CPM, per instructions given in \texttt{ProposalPrompt}.
    \item \textbf{Evaluate candidate concepts}: For each observation, the AI agent reviews the clinical notes and extracts the value of each candidate concept per instructions given in \texttt{ExtractionPrompt}.
    For each candidate concept, the AI agent fits a CPM on the training data and evaluates its performance on the validation data. The candidate concept with the best-performing CPM is selected.
    \item \textbf{Iterate}: Repeat steps 2-4 until convergence or until a maximum number of iterations is reached.
\end{enumerate}

The clinical AI team provides feedback to the AI agent by refining the hyperparameters of this procedure.
The team's key lever is through modification of free-text prompts given as instructions to the AI agent, which has the advantage of being readable by any human, even those with limited AI expertise; a default set of prompts is provided to initialize the process.
Nevertheless, the team has access to other levers as well, such as modifying which observations are included in the data, developing and/or adding new statistical tools for the AI agent to use, and deciding how samples are weighted.
To help teams come up with feedback for each round, a Protected Health Information (PHI)-compliant web interface is provided so that all team members can review the AI agent's results (\cref{fig:web_interface}).
Through this review process, the clinical AI team can decide how many rounds of HACHI to run; the following case studies found 3-4 rounds to be sufficient.

Below, we describe how clinical AI teams leveraged HACHI to learn CPMs for two prediction tasks at UCSF: whether a child presenting to the Emergency Department after head trauma will be diagnosed with traumatic brain injury (TBI) and whether an adult preparing for general surgery will develop acute kidney injury (AKI) post-surgery.

\subsection{Case Study 1: Traumatic Brain Injury (TBI)}

\begin{table}
\caption{\textbf{Description of HACHI rounds for TBI.}
\textcolor{purple}{Purple} indicates features that raised concerns. Abbreviations --- LOC: Loss of Conciousness, GCS: Glasgow Coma Scale.}
\begin{tabular}{>{\raggedright\arraybackslash}p{1cm}>{\raggedright\arraybackslash}p{4.4cm}>{\raggedright\arraybackslash}p{4cm}>{\raggedright\arraybackslash}p{4.3cm}}
\toprule
\textbf{Round} & \textbf{Key Feedback} & \textbf{Changes Made} & \textbf{Learned Concepts} \\
\midrule
\textbf{1}
& Initial exploration with default prompts
& None (baseline)
& LOC (2.70), \textcolor{purple}{Brain bleed} (1.74), Neurological event (1.58), \textcolor{purple}{Note mentions GCS} (1.23), Seizure-free (1.10)
\\
\midrule
\textbf{2}
& (1) Some concepts reflect note-writing style instead of patient characteristics; (2) Brain bleed concept suggest data leakage, investigation revealed patients with existing CT results from prior facilities
& (1) Require concept questions to have prefix ``Does the note mention the patient having...''; (2) Removed cases with existing TBI diagnosis or mentioned CT results
& LOC (3.46), Normal GCS \& age $\geq$2yr (\textcolor{purple}{1.50}), Convulsions (1.04), Altered mental status (0.82), Intact cranial nerves ($-$0.65)
\\
\midrule
\textbf{3}
& Some coefficients contradict clinical intuition
& Modified greedy concept selection to require sign of estimated coefficient to match LLM's clinical prior
& LOC (4.34), History of mild TBI (1.41), Occipital hematoma (1.21), Vision changes (1.07), Memory intact ($-$0.53)
\\
\midrule
\textbf{4}
& Model shows poor generalizability across Oakland and Mission Bay campuses due to imbalanced representation
& Introduced sample weights to weight the two campuses equally
& LOC (1.58), Altered mental status (0.73), Headache (0.25), Head trauma (0.08), Normal gait ($-$0.22)
\\
\bottomrule
\end{tabular}
\label{tab:tbi_rounds}
\end{table}

TBI is a leading cause of morbidity and mortality in children presenting to the emergency department (ED) following head trauma~\citep{langlois2006traumatic,coronado2011surveillance}.
As such, it is critical to identify which children require computed tomography (CT) imaging to avoid missing TBI cases; on the other hand, ionizing radiation from CT scans can cause lethal malignancies~\cite{brenner2002estimating,brenner2007computed}.
To guide CT imaging decisions in pediatric head trauma, the Pediatric Emergency Care Applied Research Network (PECARN) rule is currently the most widely used CPM for identifying patients at high risk of TBI~\cite{kuppermann2009identification}.
PECARN incorporates clinical signs and symptoms such as altered mental status, loss of consciousness, severe mechanism of injury, and physical examination findings (e.g., scalp hematoma, skull fractures).
While the PECARN rule is designed to achieve very high sensitivity to minimize missed TBIs, its specificity remains relatively low, potentially exposing patients to unnecessary harmful radiation~\cite{easter2014comparison,holmes2024pecarn}.

To determine whether a better-performing CPM for TBI could be learned, a clinical AI team was assembled,  including one pediatric ED clinician (A. Kornblith), an emergency clinical data analyst (N.A.), and three data scientists (J.F., C.S., A. Kothari).
Given that PECARN includes only a small number of concepts, the team decided to learn a $5$-concept CPM.
The team collated a retrospective case-control dataset with 400 cases and 400 controls from all encounters with documented ICD codes for head trauma or traumatic brain injury (TBI) diagnoses at two academic pediatric EDs within the Benioff Children’s Hospital (Oakland and Mission Bay Campuses) between March 1, 2014, and December 31, 2024.
ED Triage Notes, ED Provider Notes, and Nursing Notes were included in the analysis.
To minimize the chance of data leakage, only the History \& Physical section of ED provider notes were included and, for the other note types, only those with timestamps prior to the encounter's CT scan were included if a CT scan was performed.
Four rounds of HACHI were completed, with Table~\ref{tab:tbi_rounds} showing the learned concepts, feedback, and modifications from each round.
The full set of prompts used in each round are provided in the Supplementary Materials.

\textbf{Round 1 $\rightarrow$ 2: Data leakage and spurious correlations.} \texttt{Round 1} was run with the default template of prompts.
To the surprise of the clinical AI team, this CPM already achieved a high AUC of 0.92.
While this performance could be partly explained by the CPM's overlap in concepts with PECARN (e.g., whether the patient had experienced loss of consciousness (LOC) and having a neurological event), the team also noted two major issues that could be leading to inflated performance.
First, the AI agent had learned the concept ``whether a note mentions Glasgow Coma Scale (GCS),'' which meant that the AI agent had learned to rely on spurious correlations between the note writing style and TBI diagnosis.
Second, the AI agent identified that having a ``brain bleed'' substantially increased the risk of TBI, but brain bleeds (i.e., intracranial hemorrhages) are typically known only \textit{after} a patient has CT scan results.
After investigating this issue of data leakage, the team found that a significant proportion of patients were transferred from another ED with an existing diagnosis of TBI.
So for \texttt{Round 2}, (1) the AI agent was instructed to define concept questions using the prefix ``Does the note mention the patient having...'', to ensure that concepts extract attributes of the patient and not note writing style, and (2) all patients who already had prior CT scan results were removed from the dataset, resulting in 304 remaining cases.

\textbf{Round 2 $\rightarrow$ 3: Directional alignment with clinical prior.} \texttt{Round 2}'s CPM had a slightly lower AUC of 0.90, which was expected since Round 1's AUC was inflated.
While all the learned concepts were clinically relevant, the new concern raised by the clinical AI team was that some of the learned coefficients did not match clinical intuition.
For instance, having a normal GCS had a positive coefficient (i.e., increased risk), contrary to clinical expectations.
After discussion with the clinical AI team, the data scientists suggested modifying the greedy concept selection process in \texttt{Round 3}: rather than only selecting the candidate concept based on AUC, selection additionally required the coefficient's sign to match the LLM's clinical prior.

\textbf{Round 3 $\rightarrow$ 4: Model generalizability and fairness.} 
With the added changes, \texttt{Round 3}'s CPM simultaneously had clinically reasonable concepts, directionally aligned coefficients, and maintained the same AUC of 0.9.
As such, the clinical AI team decided to run one last round, where the team assessed the generalizability of the model across the two Oakland and Mission Bay campuses.
This revealed a surprising gap in performance: the AUCs at the Oakland and Mission Bay campuses were 0.93 and 0.71, respectively.
Investigating further, the team noted that the ratio of patients from Oakland to Mission Bay in the dataset was 3:1 and that the top feature learned by the procedure, LOC, was highly discriminative for patients at the Oakland campus but not at Mission Bay.
This was clinically plausible, as the Oakland campus functions as a safety-net, pediatric Level 1 trauma referral center that sees higher-acuity and more selectively referred children, whereas the Mission Bay campus serves as a quaternary care hospital that sees a broader population.
So for \texttt{Round 4}, the team decided to update the learning procedure to allow for sample weights, so that the two campuses could be equally weighted.
\texttt{Round 4}'s CPM not only achieved a better overall AUC of 0.91, but also better campus-specific AUCs of 0.93 and 0.80 at Oakland and Mission Bay, respectively.
The concepts in this final model are LOC, altered mental status, headache, head trauma, and having a normal gait.
The concept ``head trauma'' was initially surprising, but analyzing the LLM's annotations revealed that in real-world scenarios, there are a small percentage of cases (5\%) where there is a suspicion of unwitnessed head trauma but no reliable patient history is available (e.g., abandoned child, unclear mechanism of injury, no chief complaint on file).

\textbf{Comparator methods.}
As comparison, we evaluated the performance of PECARN by extracting relevant attributes from the same clinical notes.
In addition, we simulated the common approach to learning CPMs where a clinical expert outputs a list of potential risk factors and protective factors for inclusion in a CPM, but does not further refine this list.
To simulate this one-time brainstorming of clinical factors, we collated features that were originally considered for inclusion to the PECARN model \citep{Yen2013-xq} as well as features brainstormed by OpenEvidence \citep{Hurt2025-gc}.
Nevertheless, the AUCs of these models were 0.75 and 0.88, respectively, which were substantially worse than that achieved by HACHI.
Finally, because PECARN was designed to optimize sensitivity, we compare the specificities of the CPMs from the different methods holding sensitivity constant in Table~\ref{tab:tbi_specificity} of the Appendix, where we again find that the final model from HACHI outperforms the comparator methods.

\FloatBarrier
\subsection{Case Study 2: Acute Kidney Injury (AKI)}


    

\begin{table}
\caption{Summary of HACHI rounds for AKI. 
\blue{Blue} indicates concepts learned in response to feedback from the clinical AI team. CKD: Chronic Kidney Disease.}
\begin{tabular}{>{\raggedright\arraybackslash}p{1cm}>{\raggedright\arraybackslash}p{4cm}>{\raggedright\arraybackslash}p{4cm}>{\raggedright\arraybackslash}p{5.5cm}}
\toprule
\textbf{Round} & \textbf{Key Feedback} & \textbf{Changes Made} & \textbf{Learned Concepts} \\
\midrule
\textbf{1}
& Initial exploration with default prompts focused on patient characteristics
& None (baseline)
& Leukocytosis (1.31), Abdominal distention (0.87), Cardiac dysfunction (0.74), Tachycardia (0.71), Swelling (0.65), Systemic infection (0.57), Renal impairment (0.53), Diabetes mellitus (0.47), No kidney disease ($-$0.36), Low hematocrit ($-$1.01)
 \\
\midrule
\textbf{2}
& (1) Prompts only considered patient characteristics, but surgery type is often more influential for AKI risk; (2) Learned concepts are too broad and should be more specific; (3) clinician brainstormed additional factors such as infection and need for blood transfusion
& (1) Updated all prompts to include both patient and surgical risk factors; (2) Updated prompts to encourage more specific concepts, such as focusing on higher-risk factors; (3) Added clinician-suggested concepts as in-context learning examples in prompts
& CKD (1.28), \blue{Exploratory laparotomy} (1.03), \blue{Sepsis} (0.98), Fluid retention (0.89), Respiratory condition (0.47), \blue{Need blood transfusion during surgery} (0.43), \blue{High-risk surgery} (0.40), Heart disease (0.37), Malignancy (0.31), \blue{Minimally invasive surgery} ($-$1.01)
\\
\midrule
\textbf{3}
& (1) Concepts too vaguely defined, which may lead to inconsistent extraction between clinicians; (2) Medication concepts missing
& (1) Required concept questions to include precise definitions with examples; (2) Added medications to brainstorming prompts
& CKD (1.05), Major/high-risk surgery \blue{e.g., major abdominal surgery or anticipated blood loss $>$ 500mL} (0.91), Tachycardia \blue{$>$100 bpm} or palpitations (0.88), Active systemic infection \blue{e.g., sepsis, bacteremia, or requiring therapeutic antibiotics }(0.72), Heart failure (0.64), Sleep apnea (0.51), Urgent/emergent surgery (0.50), Obesity (0.39), Hypertension (0.31), Minimally invasive surgery \blue{e.g., laparoscopic or robotic-assisted procedures} ($-$0.74)
 \\
\bottomrule
\end{tabular}
\label{tab:aki_rounds}
\end{table}


AKI is a common but serious postoperative complication, occurring in 2–25\% of patients. When it occurs, AKI increases associated mortality and costs by two- to fivefold~\citep{wijeysundera2006improving,thakar2005clinical,chertow1997preoperative}.
Early identification of high-risk patients enables preventive measures such as optimizing hemodynamics, avoiding nephrotoxic medications, and closer monitoring.
While there are multiple well-known AKI prediction models for cardiac surgeries, there are far fewer for predicting AKI risk for lower-risk general surgeries.
The most well-known interpretable AKI prediction model for General Surgery is the Kheterpal model~\citep{Kheterpal2009-ez}, whose predictors include patient characteristics, medications, and surgery type.
Nevertheless, use of this model is limited in clinical practice, since its performance is highly dependent on the patient case mix.

To determine if a better-performing CPM for AKI could be learned, a clinical AI team was assembled with an anesthesiologist (A.B.), a clinical data analyst (J.Y.), and two data scientists (J.F. and A. Kothari).
A retrospective case-control dataset was selected uniformly at random from all General Surgery patients between January 2016 and March 2024, with 800 cases and 800 controls.
The outcome is defined as whether AKI develops within 7 days following surgery per the Kidney Disease Improving Global Outcomes (KDIGO) criteria (stage 1 or higher) \citep{Kellum2012-lm}.
To capture preoperative patient characteristics, the dataset consists of preoperative anesthesia notes.
Given that the Kheterpal model~\citep{Kheterpal2009-ez} used 11 clinical features, the team decided that a 10-concept CPM would be appropriate.
Three rounds of HACHI were completed, as shown in \cref{tab:aki_rounds}.
The full set of prompts used in each round are provided in the Supplementary Materials.

\textbf{Round 1 $\rightarrow$ 2: Missing concept category.} \texttt{Round 1} was run with the default template of prompts for learning patient characteristics associated with the prediction target.
This CPM achieved an AUC of 0.60 and identified many factors known to be associated with AKI that had been previously identified in \citep{Kheterpal2009-ez}, such as renal impairment and diabetes mellitus.
The method also identified features not in \citep{Kheterpal2009-ez}, including hematocrit levels, leukocytosis, and cardiac dysfunction.
However, the clinical AI team noted that the default prompts had directed the AI agent to only consider patient characteristics, when surgery type is often even more influential to a patient's AKI risk.
So for \texttt{Round 2}, all the LLM prompts were updated to instruct the AI agent to consider both patient and surgical risk factors.
In addition, the clinician suggested potential risk factors that could be provided to the AI agent as in-context learning examples, such as a patient's functional capacity, surgical duration, and whether the patient was coming from the ICU.

\textbf{Round 2 $\rightarrow$ 3: Increased precision.} With the inclusion of surgical risk factors, \texttt{Round 2}'s CPM had a substantially higher AUC of 0.70.
Some of these factors are already known (e.g., chronic kidney disease, fluid retention) and others are known but not included in commonly-used CPMs (e.g., minimally invasive surgery, active infection requiring antibiotics, requiring blood transfusion during surgery).
While the performance increase was promising, the clinical AI team discussed potential hurdles when implementing this CPM in practice.
The major concern was that some of the CPM concepts were too vaguely defined, so that concept extractions for the same patient may differ between clinicians, reducing reliability in practice.
So for \texttt{Round 3}, the AI agent was instructed to define concept questions more precisely using the prefix ``Does the note mention the patient having ..., e.g.,....?'' and likewise for surgery.
For instance, a valid question could be ``Does the note mention the patient having good exercise tolerance, e.g., having at least 5 METS of functional capacity or ability to climb two flights of stairs?''
The clinician also suspected that home medications may also be predictive of AKI, so the AI agent was also encouraged to consider inclusion of medication keyphrases and concepts.
The CPM in \texttt{Round 3}, the final round, achieved the highest AUC of 0.73 on the internal validation set (Period 1), suggesting that the increased precision reduced variability of the extracted concepts and thereby increased performance.
Critically, the concepts now had construct validity since they were clearly defined.
To assess temporal generalizability, the clinical AI team assembled a second test dataset from the disjoint time period of April-December 2024 (Period 2).
The AUCs in Period 2 similarly improved from Round 1 in HACHI, starting from 0.61 in Round 1 to 0.77 and 0.76 in Rounds 2 and 3, respectively.

\textbf{Comparator methods.}
As comparison, we evaluated the unweighted and weighted Kheterpal models~\citep{Kheterpal2009-ez} in two different ways: using LLM extractions from the preoperative clinical notes and using tabular data from the EHR.
The models using LLM extractions achieved AUCs of 0.65-0.66 and those using tabular data achieved AUCs of 0.64.
In addition, we simulated a clinical expert who brainstormed predictors for AKI risk without an iterative refinement process by asking OpenEvidence \citep{Hurt2025-gc} to generate a list of potential predictors.
Using the 30 concepts proposed through this approach, the resulting model achieved an AUC of 0.70.
On the temporally disjoint test dataset, the models from \citep{Kheterpal2009-ez} and the single-round brainstorming approach had similarly low AUCs of 0.67.

\section{Discussion}

This work demonstrates that meaningful collaboration between clinical AI teams and AI agents can produce high-performing CPMs that are simple to use, fully interpretable, and clinically credible.
Across two clinical prediction tasks---traumatic brain injury in children and acute kidney injury in surgical patients---we found that the HACHI co-design framework consistently improved model performance, clinical relevance, and generalizability compared to existing clinical decision instruments and approaches that learn CPMs in a zero-shot fashion.
Compared to the traditional process of training a CPM (which includes time spent on manual feature engineering, chart review, and iterative model refinement), HACHI constitutes a more efficient use of clinical AI teams: the teams in both case studies spent approximately 1-2 hours reviewing results and providing feedback per round, with 3-4 rounds sufficient to reach a satisfactory model.

These case studies highlight how human oversight is essential not only for validating AI outputs but also for shaping the model development process in ways that the AI agent could not achieve in isolation.
In the TBI case study, the clinical AI team identified critical issues that would have gone undetected in a purely automated pipeline: data leakage from transferred patients with prior CT results, spurious associations between documentation style and outcomes, and performance disparities across clinical sites with different patient populations and care pathways.
These insights led to modifications of the AI-guided CPM learning procedure not only in terms of the prompts given (e.g., constraining concept definitions to patient characteristics rather than note features and asking the AI agent whether the concepts are expected to be risk factors versus protective factors) but also the data analyzed and how the model was evaluated (e.g., removing transferred patients and reweighting the training/validation data).
Similarly, in the AKI case study, the AI agent initially focused exclusively on patient characteristics and learned concepts that were too vague for clinicians to implement in practice.
The clinical AI team had to prompt the AI agent to additionally consider other concepts such as surgical factors and had to redefine how concept questions were formulated.
Both case studies demonstrate how the clinical AI team must iteratively work with the AI agent to overcome the gulf of envisionment.

Through this collaborative process between the AI agent and clinical AI team, HACHI was able to uncover predictors commonly omitted from CPMs, while recovering commonly used predictors.
In the TBI case study, the final model learned by HACHI is a much simpler form of the PECARN model, using only a subset of features such as LOC and altered mental status, alongside the more novel feature of ``whether the patient has a normal gait.''
In the AKI case study, the final model identified multiple factors that have not been commonly included in interpretable AKI risk scores: minimally invasive surgery, tachycardia, and sleep apnea.
The more novel factors uncovered in this study may be useful more generally, although additional validation is necessary.

HACHI is designed to be simple to use, in light of recent concerns that a global CPM is unlikely to work well for all patient populations, given the complex nature of medicine \citep{Futoma2020-rc}.
Instead, these works suggest training local CPMs for local populations.
To address this concern, HACHI has minimal requirements.
It only needs access to (i) clinical notes, which are generally easy to extract from the EHR, (ii) a PHI-compliant LLM (e.g., API endpoints in the cloud), and (iii) guidance from a clinical AI team. Furthermore, the deployment of HACHI learned CPMs is straightforward.
Rather than requiring LLMs in clinical practice, clinicians can instead annotate the questions during a patient encounter.
The code for running HACHI is open-source and ready-to-use (see Code Availability section).
This allows teams to run HACHI for local patient populations, where it can discover a parsimonious set of concepts that are highly predictive locally.

\method builds on recent work using LLMs for concept bottleneck models~\citep{Oikarinen2023-ee,Patel2023-dp,McInerney2023-ab,benara2025crafting,kim2024constructing} and iterative concept refinement~\citep{ludan2023interpretable,feng2024bayesian,sun2025a,kim2024constructing}.
The most closely related method is BC-LLM~\cite{feng2024bayesian}, which uses an LLM-in-the-loop rather than humans to iteratively propose and evaluate concepts.
\method also relates to broader frameworks that combine LLMs with automatic verification loops~\citep{romera2024mathematical,novikov2025alphaevolve,singh2022explaining}, that incorporate human feedback to improve machine learning models~\cite{monarch2021human,mosqueira2023human,lage2020learning,brewster2025evaluating, Sivaraman2025-it} or that improve human-model interaction for difficult tasks~\cite{gao2024taxonomy,wucollabllm,chauhan2023interactive}.
Distinct from these prior works, HACHI integrates human domain expertise throughout the model development process, enabling clinical AI teams to identify data quality issues, enforce clinical consistency, and ensure practical usability---aspects that purely automated approaches cannot adequately address.
While we use lasso-penalized logistic regression as the underlying statistical model here, HACHI can easily be customized to use models that better fit an application, e.g. sparse integer linear models~\cite{ustun2016supersparse},
    bayesian rule lists~\cite{yang2017scalable},
    concept decision trees~\cite{grari2025act,singh2023tree,ragkousis2024tree},
    and other interpretable models~\cite{Rudin2019-yb,singh2021imodels}.

Several challenges remain before HACHI-derived CPMs can be deployed in clinical practice.
Most critically, models developed retrospectively must undergo prospective validation to ensure they perform as expected when used in real-time clinical workflows. Nevertheless, the simplicity and interpretability of HACHI-derived CPMs may facilitate more rapid prospective validation compared to complex black-box models.
Additionally, while LLM-based concept extraction is scalable, it still requires careful prompt engineering and quality control.
In our studies, we found that the iterative review process naturally led to more precise concept definitions---the clinical AI team's feedback helped refine vague concepts like ``protective factors'' into precise questions with concrete examples, which substantially improved extraction consistency.
Finally, as with any prediction model, careful attention to fairness, bias, and equity is essential, particularly when models are trained on data from specific institutions or populations that may not reflect the broader patient population.

Several limitations warrant consideration.
Most importantly, the models in this study were developed and evaluated retrospectively using historical clinical data.
While we assessed temporal generalizability using held-out time periods, retrospective evaluation may not fully capture how models will perform when deployed prospectively in real-time clinical workflows, so prospective validation is still needed.
In addition, though these case studies involved multiple sites, the analyses and models were conducted at a single academic medical center with specific documentation practices and patient populations.
The team members applying HACHI are also from this same institution, and it is to be determined if a different clinical AI team would have achieved similar performance and learned similar concepts.
Finally, the quality of HACHI-derived models depends on the LLM's ability to accurately extract concepts from clinical notes and the types of clinical notes that are included.
While prior work has found that GPT models are highly accurate at extracting concepts from medical texts \citep{Agrawal2022-me,kornblith2025analyzing}, LLMs differ in their extraction accuracy and, consequently, the CPM's performance may also differ.

Avenues for future research include extending HACHI's applicability to more complex clinical scenarios by incorporating multimodal data such as imaging, laboratory results, or time series data, as has been done in prior CBM works~\cite{shi2025multimodal,pang2024integrating}.
In addition, other frameworks may be explored for incorporating human feedback into the co-design process, such as ways to reduce the burden on clinical experts while maintaining quality.
Finally, although HACHI was tested in two clinical examples, the framework is applicable more broadly, particularly to fields beyond healthcare that also have large amounts of unstructured data.
For this broader framework, perhaps a more suitable name is HACHII, ``\underline{H}uman+\underline{A}gent \underline{C}o-design for \underline{H}ighly \underline{I}nterpretable \underline{I}nstruments.''

\section{Methods}
\label{sec:methods}
\FloatBarrier

\subsection{AI agentic CPM learning procedure}

The AI agent is an encapsulated large language model (LLM) managed entirely through text-based prompts and outputs.
The agent conducts a greedy hill-climbing procedure to select $k$ concepts that are most predictive of the prediction target, subject to constraints specified by the clinical AI team.
The procedure can be broken into the three main steps highlighted in \cref{fig:intro}a, beginning with the initialization in Step 1.



\paragraph{Step 1: Initialize the CPM} 
To brainstorm possible concepts, HACHI iterates through the clinical notes in the dataset, appending each note to a \texttt{KeyphrasePrompt}. 
This prompt is used to instruct the LLM to extract a list of `keyphrases' or `keywords' that represent the content of the note.
The data is then split into a training and validation partition.
Applying a statistical tool (e.g. ridge-penalized logistic regression), HACHI fits a bag-of-words model on the training partition to identify keyphrases that are most associated with the outcome of interest.
Next, to initialize the concepts, HACHI appends the top keyphrases to a \texttt{ConceptInitializationProposalPrompt}.
This prompt instructs the LLM to make use of these keyphrases, together with its world knowledge, to propose $k$ candidate ``concept questions''.
Concept questions are defined as human-interpretable yes/no questions, e.g. ``Does this note mention the patient having a history of smoking?'' Concept questions can encompass multiple keyphrases (i.e., the concepts may be hierarchical).

To convert the extracted concepts into a CPM, the $k$ candidate ``concept questions'' are first extracted from the LLM's response.
For each question, HACHI iterates through the clinical notes, appending both the question and the note to a \texttt{ConceptAnnotationPrompt}, which instructs the LLM to give a yes/no response to the question posed regarding the note.
Finally, HACHI fits a $k$-concept CPM on the training partition.
 
Now that the CPM has been initialized, HACHI cyclically traverses the concept positions $j=1,2,\ldots,k$, evaluating each current concept for potential replacement by a more predictive concept. For each concept position, it conducts steps 2 and 3 as follows:

\begin{itemize}
    \item[] \textbf{Step 2: Propose candidate concepts.} The AI agent fits a bag-of-words model on the training partition to identify keyphrases that are most associated with the outcome of interest $Y$, but this time adjusting for existing concepts that are not up for replacement.
    HACHI appends the top keyphrases to a \texttt{ConceptReplacementProposalPrompt}, which instructs the LLM to make use of these keyphrases, together with its world knowledge, to propose $m$ candidate replacement ``concept questions''.
    The $m$ candidate ``concept questions'' are extracted from the LLM's response.
    \item[] \textbf{Step 3: Evaluate candidate concepts.} For each question, the AI agent iterates through the clinical notes, appending both the question and the note to a \texttt{ConceptAnnotationPrompt}, which instructs the LLM to give a yes/no response to the question posed regarding the note.
    The resulting annotations are then used by the AI agent to 
    fit
    a $k$-concept CPM (again using lasso-penalized logistic regression) on the training partition for each candidate concept by combining the candidate with the existing $k-1$ concepts.
    Applying a statistical validation metric (e.g. AUC), HACHI evaluates the resulting CPM's performance on the validation partition.
    If the best-performing candidate concept outperforms the existing concept being considered for replacement, HACHI replaces it with this best-performing candidate.
\end{itemize}

\paragraph{Implementation details}
In our experiments, we performed 10 iterations of the AI agent loop consisting of Steps 2 and 3. 
In ablation studies, we did not find significant improvement in the validation AUC with more iterations.

The AI-guided CPM learning procedure is stochastic, due to both sample splitting and inclusion of the AI agent.
Furthermore, there is inherent uncertainty regarding the true concepts that are relevant for predicting the outcome of interest, due to finite sample sizes and the infinite number of possible concepts.
To address both concerns, we run the AI-guided CPM learning procedure for multiple seeds in parallel.
Differences between the learned CPMs can reveal uncertainty in the truly relevant concepts and stability of the learning process, enabling more efficient Human-AI interaction.

\subsection{Human-AI interaction framework}

\paragraph{Interface for reviewing results from the AI agent}

\begin{figure}
    \centering
    \includegraphics[width=1.0\linewidth]{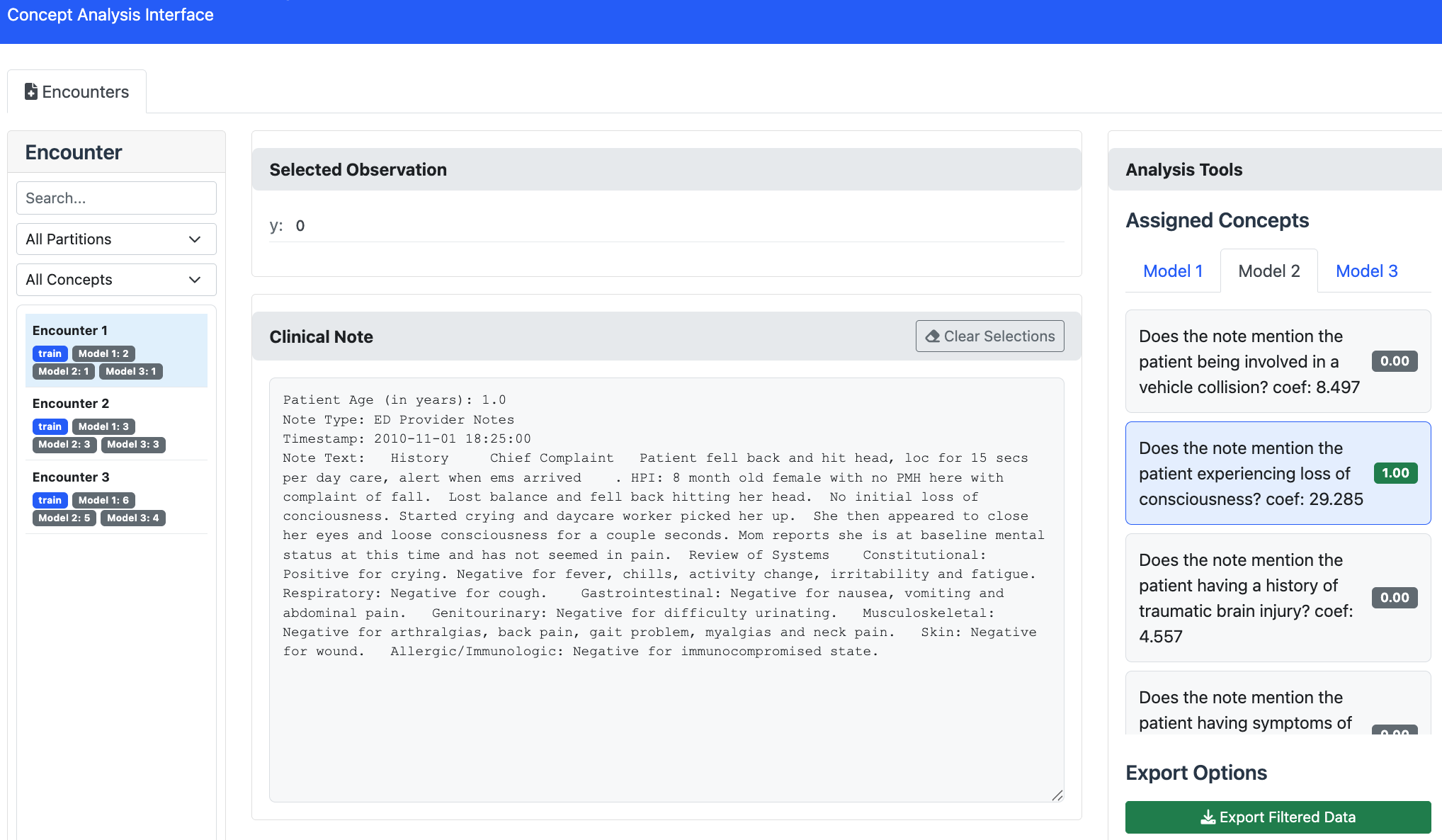}
    \caption{\textbf{PHI-compliant web interface for auditing the AI-agent CPM learning procedure}. It shows (i) the learned factors, (ii) the LLM concept annotations at a patient/note level, (iii) which patients received an incorrect prediction, and (iv) the performance of the CPM on a held-out test set.}
    \label{fig:web_interface}
\end{figure}

To help the human expert determine how the prompts should be modified each round, we created an interactive PHI-compliant single-page locally-hosted webpage to show:
(i) the learned factors, (ii) the LLM concept annotations at a patient/note level, (iii) which patients received an incorrect prediction, and (iv) the performance of the CPM on a held-out test set;
see \cref{fig:web_interface} for more details.
A meeting between the data scientists and clinical experts is also held to review the learned CPMs and brainstorm ways to improve it.
Data scientists then translate feedback from this meeting into actual LLM prompts and/or code updates.

\paragraph{Human-in-the-loop control and prompt steering}

The clinical AI team exerts control over the AI agent primarily by modifying the four prompts: 
\texttt{KeyphrasePrompt},
\texttt{ConceptInitializationProposalPrompt},
\texttt{ConceptReplacementProposalPrompt}, and
\texttt{ConceptAnnotationPrompt}.
However, the team may also opt for other modifications as well, such as the composition of the dataset, modifications to the statistical tools used by the AI agent (e.g. the extension of the tool to allow for sample weights), and modifications to the code for how candidate concepts are selected in the optimization process.
More complex modifications may require code-based changes to the AI-guided CPM learning procedure, rather than simply modifying the inputs to the procedure.

\subsection{Dataset extended details}


\paragraph{Data collection for traumatic brain injury (TBI)}
To ensure a comprehensive capture of relevant cases, we included any encounter with documented head trauma or a traumatic brain injury (TBI) diagnosis, as coded by the following ICD-10 (International Classification of Diseases, Tenth Revision [ICD-10]) codes: S06, S00.03, S00.83, S00.93, S02.0, S02.1, S02.80, S02.81, S02.91, S07.1, S07.8, S07.9, S09.8, and S09.90 for head trauma, and S06.1, S06.2, S06.3, S06.5, S06.6, S06.89, S06.9, S04.02, S04.03, S04.04, S07.1, and T74.4 for TBI. Encounters with at least one qualifying hospital-wide diagnostic code within the same encounter or occurring up to one calendar day before or after the ED visit were included, to capture both ED and Inpatient Diagnoses. 
Concatenated, the notes per encounter contain an average of 1,454 tokens.
\cref{tab:prevalence} (top) shows the prevalence of various concepts in the final dataset.



\paragraph{Data collection for acute kidney injury (AKI)}
The retrospective case-control cohort was assembled from all General Surgery cases  between January 2016 to March 2024 at UCSF.
Inclusion criteria included adult patients undergoing inpatient procedures (including emergency surgeries) with at least one serum creatinine (sCr) value in the 90 days preceding surgery and  at least one serum sCr in the 7 days following surgery.
The main outcome was AKI according to the KDIGO criteria in the 7 days following surgery: [48 h maximum postoperative
sCr] - [last preoperative sCr] $\ge$ 0.3 mg/dL or [7 day maximum postoperative sCr]/[last preoperative sCr] $\ge$ 1.5 \citep{Kellum2012-lm}.
We analyze the preoperative note written by the anesthesiologist, which has an average of 1991 tokens.
\cref{tab:prevalence} (bottom) shows the prevalence of various concepts in the final dataset.
Among the AKI cases, the prevalence of AKI Stage 1 was 0.74 (0.66, 0.80), AKI Stage 2 was 0.12 (0.08, 0.14), and AKI Stage 3 was 0.16 (0.12, 0.20).


\begin{table}
    \caption{Final concepts in CPM for AKI and TBI learned by HACHI. Coefficients and prevalences shown.}

\begin{tabular}{p{10cm}cc}
\toprule
TBI Concept & Coefficient & Prevalence \\
\midrule
... the patient having normal gait? & -0.22 & 0.56 (0.51, 0.61) \\
... the patient experiencing head trauma? & 0.08 & 0.95 (0.93, 0.97) \\
... the patient experiencing a headache? & 0.25 & 0.42 (0.37, 0.47) \\
... the patient having altered mental status? & 0.73 & 0.13 (0.10, 0.16) \\
... the patient experiencing unconsciousness? & 1.58 & 0.39 (0.34, 0.44) \\
\toprule
AKI Concept & Coefficient & Prevalence \\
\midrule
... the surgery being minimally invasive, e.g., laparoscopic or robotic-assisted procedures? & -0.74 & 0.41 (0.37, 0.44) \\
... the patient having hypertension, e.g., high blood pressure or antihypertensive medication use? & 0.31 & 0.54 (0.51, 0.58) \\
... the patient having obesity, e.g., high BMI or overweight status? & 0.39 & 0.32 (0.29, 0.36) \\
... the surgery being urgent or emergent, e.g., requiring immediate attention or intervention? & 0.50 & 0.21 (0.18, 0.23) \\
... the patient having a history of sleep apnea, e.g., witnessed apnea or CPAP use? & 0.51 & 0.20 (0.17, 0.23) \\
... the patient having heart failure, e.g., reduced ejection fraction or history of heart failure exacerbations? & 0.64 & 0.08 (0.06, 0.10) \\
... the patient having an active systemic infection, e.g., sepsis, bacteremia, or requiring therapeutic antibiotics? & 0.72 & 0.07 (0.05, 0.09) \\
... the patient having tachycardia, e.g., heart rate $>$ 100 bpm or palpitations? & 0.88 & 0.16 (0.14, 0.19) \\
... the surgery being major or high-risk, e.g., major abdominal surgery or anticipated blood loss $>$ 500mL? & 0.91 & 0.40 (0.36, 0.43) \\
... the patient having chronic kidney disease, e.g., a history of CKD or elevated creatinine levels? & 1.05 & 0.13 (0.11, 0.16) \\
\bottomrule
\end{tabular}

    \label{tab:prevalence}
\end{table}

\FloatBarrier
\section{Declaration Statements}
\subsection{Data Availability}
Data that support the findings of this study are not publicly available due to the use of protected health information. Deidentified data can be made available from the corresponding author upon reasonable request.
 
\subsection{Code Availability}
An open-source python package for the HACHI framework is publicly available at \url{http://github.com/jjfenglab/HACHI}, which includes code for reproducing this work, web interfaces for reviewing outputs from the AI agent, and tutorials for running the AI-guided CPM learning procedure.
 
\subsection{Acknowledgements}
J.F. and P.V. were supported through a Patient-Centered Outcomes Research Institute\textsuperscript{®} (PCORI\textsuperscript{®}) Award (ME-2022C1-25619), A.B. was supported by National Institute of General Medical Sciences of the National Institutes of Health (K23GM151611), and A. Kornblith was supported by Eunice Kennedy Shriver National Institute of Child Health and Human Development of the National Institutes of Health (K23HD110716) for this project.
The views presented in this work are solely the responsibility of the author(s) and do not necessarily represent the views of the PCORI\textsuperscript{®}, its Board of Governors or Methodology Committee, or the National Institutes of Health.
The PROSPECT lab (J.F., L.Z., A. Kothari, P.V.) would also like to acknowledge the support from the Zuckerberg Priscilla Chan quality improvement fund via the San Francisco General Foundation for this project.
The authors thank the UCSF AI Tiger Team, Academic Research Services, Research Information Technology, and the Chancellor’s Task Force for Generative AI for their technical support related to the use of Versa API gateway.
The authors would also like to thank Joanne Yim for helping with extracting data for the AKI case study.
 
\subsection{Author Contributions}
J.F., A. Kothari, L.Z., P.V., Y.T., and C.S. developed the methodology and software.
J.F., A. Kothari, A.B., N.A., and A. Kornblith curated data.
J.F., A. Kothari, P.V., A.B., N.A., A. Kornblith, Y.T., and C.S. analyzed results and validated the methodology.
J.F., A. Kothari, Y.T., and C.S. wrote the main manuscript text.
All authors contributed to the conceptualization of the work and reviewed/edited the manuscript.
 
\subsection{Competing Interests}
The authors have no competing interests as defined by Nature Portfolio, or other interests that might be perceived to influence the results and/or discussion reported in this paper.

\FloatBarrier
{
    \bibliography{refs}

@article{novikov2025alphaevolve,
  title={AlphaEvolve: A coding agent for scientific and algorithmic discovery},
  author={Novikov, Alexander and V{\~u}, Ng{\^a}n and Eisenberger, Marvin and Dupont, Emilien and Huang, Po-Sen and Wagner, Adam Zsolt and Shirobokov, Sergey and Kozlovskii, Borislav and Ruiz, Francisco JR and Mehrabian, Abbas and others},
  journal={arXiv preprint arXiv:2506.13131},
  year={2025}
}

@article{lage2020learning,
  title={Learning interpretable concept-based models with human feedback},
  author={Lage, Isaac and Doshi-Velez, Finale},
  journal={arXiv preprint arXiv:2012.02898},
  year={2020}
}

@article{ragkousis2024tree,
  title={Tree-Based Leakage Inspection and Control in Concept Bottleneck Models},
  author={Ragkousis, Angelos and Parbhoo, Sonali},
  journal={arXiv preprint arXiv:2410.06352},
  year={2024}
}

@INPROCEEDINGS{McInerney2023-ab,
  title     = "{CHiLL}: Zero-shot Custom Interpretable Feature Extraction from
               Clinical Notes with Large Language Models",
  author    = "McInerney, Denis Jered and Young, Geoffrey and van de Meent,
               Jan-Willem and Wallace, Byron C",
  booktitle = "The 2023 Conference on Empirical Methods in Natural Language
               Processing",
  month     =  dec,
  year      =  2023
}

@inproceedings{singh2023tree,
  title={Tree prompting: Efficient task adaptation without fine-tuning},
  author={Singh, Chandan and Morris, John and Rush, Alexander M and Gao, Jianfeng and Deng, Yuntian},
  booktitle={Proceedings of the 2023 Conference on Empirical Methods in Natural Language Processing},
  pages={6253--6267},
  year={2023}
}

@article{ludan2023interpretable,
  title={Interpretable-by-Design Text Classification with Iteratively Generated Concept Bottleneck},
  author={Ludan, Josh Magnus and Lyu, Qing and Yang, Yue and Dugan, Liam and Yatskar, Mark and Callison-Burch, Chris},
  journal={arXiv preprint arXiv:2310.19660},
  year={2023}
}

@article{brenner2002estimating,
  title={Estimating cancer risks from pediatric CT: going from the qualitative to the quantitative},
  author={Brenner, David J},
  journal={Pediatric radiology},
  volume={32},
  number={4},
  pages={228--231},
  year={2002},
  publisher={Springer-Verlag Berlin/Heidelberg}
}

@inproceedings{yang2017scalable,
  title={Scalable Bayesian rule lists},
  author={Yang, Hongyu and Rudin, Cynthia and Seltzer, Margo},
  booktitle={International conference on machine learning},
  pages={3921--3930},
  year={2017},
  organization={PMLR}
}

@article{ustun2016supersparse,
  title={Supersparse linear integer models for optimized medical scoring systems},
  author={Ustun, Berk and Rudin, Cynthia},
  journal={Machine Learning},
  volume={102},
  number={3},
  pages={349--391},
  year={2016},
  publisher={Springer}
}

@inproceedings{gao2024taxonomy,
  title={A taxonomy for human-llm interaction modes: An initial exploration},
  author={Gao, Jie and Gebreegziabher, Simret Araya and Choo, Kenny Tsu Wei and Li, Toby Jia-Jun and Perrault, Simon Tangi and Malone, Thomas W},
  booktitle={Extended Abstracts of the CHI Conference on Human Factors in Computing Systems},
  pages={1--11},
  year={2024}
}

@inproceedings{wucollabllm,
  title={CollabLLM: From Passive Responders to Active Collaborators},
  author={Wu, Shirley and Galley, Michel and Peng, Baolin and Cheng, Hao and Li, Gavin and Dou, Yao and Cai, Weixin and Zou, James and Leskovec, Jure and Gao, Jianfeng},
  booktitle={Forty-second International Conference on Machine Learning}
}

@article{brenner2007computed,
  title={Computed tomography—an increasing source of radiation exposure},
  author={Brenner, David J and Hall, Eric J},
  journal={New England journal of medicine},
  volume={357},
  number={22},
  pages={2277--2284},
  year={2007},
  publisher={Mass Medical Soc}
}

@inproceedings{
sun2025a,
title={A General Framework for Producing Interpretable Semantic Text Embeddings},
author={Yiqun Sun and Qiang Huang and Yixuan Tang and Anthony Kum Hoe Tung and Jun Yu},
booktitle={The Thirteenth International Conference on Learning Representations},
year={2025},
url={https://openreview.net/forum?id=23uY3FpQxc}
}

@article{coronado2011surveillance,
  title={Surveillance for traumatic brain injury-related deaths: United States, 1997-2007},
  author={Coronado, Victor G and Xu, Likang and Basavaraju, Sridhar V and McGuire, Lisa C and Wald, Marlena M and Faul, Mark D and Guzman, Bernardo R and Hemphill, John D and Centers for Disease Control and Prevention (CDC) and others},
  year={2011},
  publisher={US Department of Health and Human Services, Centers for Disease Control and~…}
}

@article{langlois2006traumatic,
  title={Traumatic brain injury in the United States: emergency department visits, hospitalizations, and deaths},
  author={Langlois, Jean A and Rutland-Brown, Wesley and Thomas, Karen E},
  year={2006}
}

@article{singh2022explaining,
  title={Explaining patterns in data with language models via interpretable autoprompting},
  author={Singh, Chandan and Morris, John X and Aneja, Jyoti and Rush, Alexander M and Gao, Jianfeng},
  journal={arXiv preprint arXiv:2210.01848},
  year={2022}
}

@article{kuppermann2009identification,
	title        = {Identification of children at very low risk of clinically-important brain injuries after head trauma: a prospective cohort study},
	author       = {Kuppermann, Nathan and Holmes, James F and Dayan, Peter S and Hoyle, John D and Atabaki, Shireen M and Holubkov, Richard and Nadel, Frances M and Monroe, David and Stanley, Rachel M and Borgialli, Dominic A and others},
	year         = 2009,
	journal      = {The Lancet},
	publisher    = {Elsevier},
	volume       = 374,
	number       = 9696,
	pages        = {1160--1170}
}

@article{singh2021imodels,
	title        = {imodels: a python package for fitting interpretable models},
	author       = {Singh, Chandan and Nasseri, Keyan and Tan, Yan Shuo and Tang, Tiffany and Yu, Bin},
	year         = 2021,
	journal      = {Journal of Open Source Software},
	publisher    = {The Open Journal},
	volume       = 6,
	number       = 61,
	pages        = 3192,
	doi          = {10.21105/joss.03192},
	url          = {https://doi.org/10.21105/joss.03192}
}

@article{kornblith2025analyzing,
  title={Analyzing patient perspectives with large language models: a cross-sectional study of sentiment and thematic classification on exception from informed consent},
  author={Kornblith, Aaron E and Singh, Chandan and Innes, Johanna C and Chang, Todd P and Adelgais, Kathleen M and Holsti, Maija and Kim, Joy and McClain, Bradford and Nishijima, Daniel K and Rodgers, Steffanie and others},
  journal={Scientific reports},
  volume={15},
  number={1},
  pages={6179},
  year={2025},
  publisher={Nature Publishing Group UK London}
}

@inproceedings{pang2024integrating,
  title={Integrating clinical knowledge into concept bottleneck models},
  author={Pang, Winnie and Ke, Xueyi and Tsutsui, Satoshi and Wen, Bihan},
  booktitle={International Conference on Medical Image Computing and Computer-Assisted Intervention},
  pages={243--253},
  year={2024},
  organization={Springer}
}

@inproceedings{shi2025multimodal,
  title={Multimodal Concept Bottleneck Models},
  author={Shi, Tongqing and Yan, Ge and Oikarinen, Tuomas and Weng, Tsui-Wei},
  booktitle={Mechanistic Interpretability Workshop at NeurIPS 2025}
}

@inproceedings{chauhan2023interactive,
  title={Interactive concept bottleneck models},
  author={Chauhan, Kushal and Tiwari, Rishabh and Freyberg, Jan and Shenoy, Pradeep and Dvijotham, Krishnamurthy},
  booktitle={Proceedings of the aaai conference on artificial intelligence},
  volume={37},
  number={5},
  pages={5948--5955},
  year={2023}
}

@article{grari2025act,
  title={ACT: Agentic Classification Tree},
  author={Grari, Vincent and Arni, Tim and Laugel, Thibault and Lamprier, Sylvain and Zou, James and Detyniecki, Marcin},
  journal={arXiv preprint arXiv:2509.26433},
  year={2025}
}

@ARTICLE{Patel2023-dp,
  title         = "Learning Interpretable Style Embeddings via Prompting {LLMs}",
  author        = "Patel, Ajay and Rao, Delip and Kothary, Ansh and McKeown,
                   Kathleen and Callison-Burch, Chris",
  journal       = "Proc. Conf. Empir. Methods Nat. Lang. Process.",
  month         =  may,
  year          =  2023,
  archivePrefix = "arXiv",
  primaryClass  = "cs.CL"
}

@ARTICLE{Oikarinen2023-ee,
  title         = "Label-Free Concept Bottleneck Models",
  author        = "Oikarinen, Tuomas and Das, Subhro and Nguyen, Lam M and Weng,
                   Tsui-Wei",
  journal       = "International Conference on Learning Representations",
  month         =  apr,
  year          =  2023,
  archivePrefix = "arXiv",
  primaryClass  = "cs.LG"
}

@INPROCEEDINGS{Ramaswamy2023-qn,
  title     = "Overlooked Factors in Concept-Based Explanations: Dataset Choice,
               Concept Learnability, and Human Capability",
  author    = "Ramaswamy, Vikram V and Kim, Sunnie S Y and Fong, Ruth and
               Russakovsky, Olga",
  booktitle = "Proceedings of the IEEE/CVF Conference on Computer Vision and
               Pattern Recognition",
  pages     = "10932--10941",
  year      =  2023
}

@article{benara2025crafting,
  title={Crafting Interpretable Embeddings for Language Neuroscience by Asking LLMs Questions},
  author={Benara, Vinamra and Singh, Chandan and Morris, John and Antonello, Richard and Stoica, Ion and Huth, Alexander and Gao, Jianfeng},
  journal={Advances in Neural Information Processing Systems},
  volume={37},
  pages={124137--124162},
  year={2025}
}

@article{easter2014comparison,
  title={Comparison of PECARN, CATCH, and CHALICE rules for children with minor head injury: a prospective cohort study},
  author={Easter, Joshua S and Bakes, Katherine and Dhaliwal, Jasmeet and Miller, Michael and Caruso, Emily and Haukoos, Jason S},
  journal={Annals of emergency medicine},
  volume={64},
  number={2},
  pages={145--152},
  year={2014},
  publisher={Elsevier}
}

@article{holmes2024pecarn,
  title={PECARN prediction rules for CT imaging of children presenting to the emergency department with blunt abdominal or minor head trauma: a multicentre prospective validation study},
  author={Holmes, James F and Yen, Kenneth and Ugalde, Irma T and Ishimine, Paul and Chaudhari, Pradip P and Atigapramoj, Nisa and Badawy, Mohamed and McCarten-Gibbs, Kevan A and Nielsen, Donovan and Sage, Allyson C and others},
  journal={The Lancet Child \& Adolescent Health},
  volume={8},
  number={5},
  pages={339--347},
  year={2024},
  publisher={Elsevier}
}

@book{monarch2021human,
  title={Human-in-the-Loop Machine Learning: Active learning and annotation for human-centered AI},
  author={Monarch, Robert Munro},
  year={2021},
  publisher={Simon and Schuster}
}

@article{mosqueira2023human,
  title={Human-in-the-loop machine learning: a state of the art},
  author={Mosqueira-Rey, Eduardo and Hern{\'a}ndez-Pereira, Elena and Alonso-R{\'\i}os, David and Bobes-Bascar{\'a}n, Jos{\'e} and Fern{\'a}ndez-Leal, {\'A}ngel},
  journal={Artificial Intelligence Review},
  volume={56},
  number={4},
  pages={3005--3054},
  year={2023},
  publisher={Springer}
}

@article{wijeysundera2006improving,
  title={Improving the identification of patients at risk of postoperative renal failure after cardiac surgery.},
  author={Wijeysundera, Duminda N and Karkouti, Keyvan and Beattie, W Scott and Rao, Vivek and Ivanov, Joan},
  journal={Anesthesiology},
  volume={104},
  number={1},
  pages={65--72},
  year={2006}
}

@article{chertow1997preoperative,
  title={Preoperative renal risk stratification},
  author={Chertow, Glenn M and Lazarus, J Michael and Christiansen, Cindy L and Cook, E Francis and Hammermeister, Karl E and Grover, Frederick and Daley, Jennifer},
  journal={Circulation},
  volume={95},
  number={4},
  pages={878--884},
  year={1997},
  publisher={Lippincott Williams \& Wilkins}
}

@article{thakar2005clinical,
  title={A clinical score to predict acute renal failure after cardiac surgery},
  author={Thakar, Charuhas V and Arrigain, Susana and Worley, Sarah and Yared, Jean-Pierre and Paganini, Emil P},
  journal={Journal of the American Society of Nephrology},
  volume={16},
  number={1},
  pages={162--168},
  year={2005},
  publisher={LWW}
}

@article{brewster2025evaluating,
  title={Evaluating human-in-the-loop strategies for artificial intelligence-enabled translation of patient discharge instructions: a multidisciplinary analysis},
  author={Brewster, Ryan CL and Tse, Gabe and Fan, Angela L and Elborki, Marwa and Newell, Maiah and Gonzalez, Priscilla and Hoq, Amitra and Chang, Crystal and Chowdhury, Maksud and Geeti, Adiba and others},
  journal={NPJ digital medicine},
  volume={8},
  number={1},
  pages={629},
  year={2025},
  publisher={Nature Publishing Group UK London}
}

@inproceedings{
    feng2024bayesian,
    title={Bayesian Concept Bottleneck Models with {LLM} Priors},
    author={Jean Feng and Avni Kothari and Lucas Zier and Chandan Singh and Yan Shuo Tan},
    booktitle={The Thirty-ninth Annual Conference on Neural Information Processing Systems},
    year={2025},
    url={https://openreview.net/forum?id=oXSkzIXgbk}
}

@article{romera2024mathematical,
  title={Mathematical discoveries from program search with large language models},
  author={Romera-Paredes, Bernardino and Barekatain, Mohammadamin and Novikov, Alexander and Balog, Matej and Kumar, M Pawan and Dupont, Emilien and Ruiz, Francisco JR and Ellenberg, Jordan S and Wang, Pengming and Fawzi, Omar and others},
  journal={Nature},
  volume={625},
  number={7995},
  pages={468--475},
  year={2024},
  publisher={Nature Publishing Group UK London}
}

@inproceedings{kim2024constructing,
  title={Constructing concept-based models to mitigate spurious correlations with minimal human effort},
  author={Kim, Jeeyung and Wang, Ze and Qiu, Qiang},
  booktitle={European Conference on Computer Vision},
  pages={137--153},
  year={2024},
  organization={Springer}
}

@ARTICLE{Kheterpal2009-ez,
  title     = "Development and validation of an acute kidney injury risk index
               for patients undergoing general surgery: results from a national
               data set: Results from a national data set",
  author    = "Kheterpal, Sachin and Tremper, Kevin K and Heung, Michael and
               Rosenberg, Andrew L and Englesbe, Michael and Shanks, Amy M and
               Campbell, Jr, Darrell A",
  journal   = "Anesthesiology",
  publisher = "Ovid Technologies (Wolters Kluwer Health)",
  volume    =  110,
  number    =  3,
  pages     = "505--515",
  month     =  mar,
  year      =  2009,
  language  = "en"
}

@ARTICLE{Yen2013-xq,
  title     = "Interobserver agreement in the clinical assessment of children
               with blunt abdominal trauma",
  author    = "Yen, Kenneth and Kuppermann, Nathan and Lillis, Kathleen and
               Monroe, David and Borgialli, Dominic and Kerrey, Benjamin T and
               Sokolove, Peter E and Ellison, Angela M and Cook, Lawrence J and
               Holmes, James F and {Intra-abdominal Injury Study Group for the
               Pediatric Emergency Care Applied Research Network (PECARN)}",
  journal   = "Acad. Emerg. Med.",
  publisher = "Wiley",
  volume    =  20,
  number    =  5,
  pages     = "426--432",
  month     =  may,
  year      =  2013,
  language  = "en"
}

@ARTICLE{Kothari2026-zo,
  title         = "When the domain expert has no time and the {LLM} developer
                   has no clinical expertise: Real-world lessons from {LLM}
                   co-design in a safety-net hospital",
  author        = "Kothari, Avni and Vossler, Patrick and Digitale, Jean and
                   Forouzannia, Mohammad and Rosenberg, Elise and Lee, Michele
                   and Bryant, Jennee and Molina, Melanie and Marks, James and
                   Zier, Lucas and Feng, Jean",
  journal       = "Proc. Conf. AAAI Artif. Intell.",
  year          =  2026,
  archivePrefix = "arXiv",
  primaryClass  = "cs.CY"
}

@INPROCEEDINGS{Subramonyam2024-at,
  title     = "Bridging the gulf of envisioning: Cognitive challenges in prompt
               based interactions with {LLMs}",
  author    = "Subramonyam, Hari and Pea, Roy and Pondoc, Christopher and
               Agrawala, Maneesh and Seifert, Colleen",
  booktitle = "Proceedings of the CHI Conference on Human Factors in Computing
               Systems",
  publisher = "ACM",
  address   = "New York, NY, USA",
  volume    =  31,
  pages     = "1--19",
  month     =  may,
  year      =  2024
}

@ARTICLE{Ranzani2025-yh,
  title     = "Development and validation of the Sequential Organ Failure
               Assessment ({SOFA})-2 score",
  author    = "Ranzani, Otavio T and Singer, Mervyn and Salluh, Jorge I F and
               Shankar-Hari, Manu and Pilcher, David and Berger-Estilita, Joana
               and Coopersmith, Craig M and Juffermans, Nicole P and Laffey,
               John and Reinikainen, Matti and Neto, Ary Serpa and Tavares,
               Miguel and Timsit, Jean-François and Arias Lopez, Maria Del Pilar
               and Arulkumaran, Nish and Aryal, Diptesh and Azoulay, Elie and
               Celi, Leo Anthony and Chaudhuri, Dipayan and De Lange, Dylan and
               De Waele, Jan and Dos Santos, Claudia C and Du, Bin and Einav,
               Sharon and Engelbrecht, Teresa and Fazla, Fathima and Ferrer,
               Ricard and Finazzi, Stefano and Fujii, Tomoko and Gershengorn,
               Hayley B and Greene, John D and Haniffa, Rashan and Hao, Sicheng
               and Hasan, Mohd Shahnaz and Hollenberg, Steve and Ippolito,
               Mariachiara and Jung, Christian and Kirov, Mikhail and Kobari,
               Shigetaka and Lakbar, Inès and Lipman, Jeffrey and Liu, Vincent
               and Liu, Xiaoli and Lobo, Suzana M and Magatti, Demetrio and
               Martin, Greg S and Metnitz, Barbara and Metnitz, Philipp and
               Myatra, Sheila N and Oczkowski, Simon and Paiva, José-Artur and
               Paruk, Fathima and Pekkarinen, Pirkka T and Piquilloud, Lise and
               Pölkki, Anssi and Prescott, Hallie C and Blaser, Annika Reintam
               and Rezende, Ederlon and Robba, Chiara and Rochwerg, Bram and
               Ruckly, Stephane and Samei, Rasoul and Schenck, Edward J and
               Secombe, Paul and Sendagire, Cornelius and Siaw-Frimpong, Moses
               and Simpkin, Andrew J and Soares, Márcio and Summers, Charlotte
               and Szczeklik, Wojciech and Takala, Jukka and Tanaka, Shiro and
               Tricella, Giovanni and Vincent, Jean-Louis and Wendon, Julia and
               Zampieri, Fernando G and Rhodes, Andrew and Moreno, Rui",
  journal   = "JAMA",
  publisher = "American Medical Association (AMA)",
  volume    =  334,
  number    =  23,
  pages     = "2090--2103",
  month     =  dec,
  year      =  2025,
  language  = "en"
}

@ARTICLE{Kellum2012-lm,
  title     = "Kidney disease: Improving global outcomes ({KDIGO}) acute kidney
               injury work group. {KDIGO} clinical practice guideline for acute
               kidney injury",
  author    = "Kellum, John A and Lameire, Norbert and Aspelin, Peter and
               Barsoum, Rashad S and Burdmann, Emmanuel A and Goldstein, Stuart
               L and Herzog, Charles A and Joannidis, Michael and Kribben,
               Andreas and Levey, Andrew S and MacLeod, Alison M and Mehta,
               Ravindra L and Murray, Patrick T and Naicker, Saraladevi and
               Opal, Steven M and Schaefer, Franz and Schetz, Miet and Uchino,
               Shigehiko",
  journal   = "Kidney Int. Suppl. (2011)",
  publisher = "Elsevier BV",
  volume    =  2,
  number    =  1,
  pages     =  1,
  month     =  mar,
  year      =  2012,
  language  = "en"
}

@ARTICLE{Obra2025-br,
  title     = "Potential for algorithmic bias in Clinical decision instrument
               development",
  author    = "Obra, Jed Keenan and Singh, Chandan and Watkins, Kenshata and
               Feng, Jean and Obermeyer, Ziad and Kornblith, Aaron",
  journal   = "NPJ Digit. Med.",
  publisher = "Springer Science and Business Media LLC",
  pages     = "1--7",
  month     =  dec,
  year      =  2025,
  language  = "en"
}

@ARTICLE{Rudin2019-yb,
  title     = "Stop explaining black box machine learning models for high stakes
               decisions and use interpretable models instead",
  author    = "Rudin, Cynthia",
  journal   = "Nature Machine Intelligence",
  publisher = "Nature Publishing Group",
  volume    =  1,
  number    =  5,
  pages     = "206--215",
  month     =  may,
  year      =  2019,
  language  = "en"
}

@ARTICLE{Koh2020-kl,
  title   = "Concept Bottleneck Models",
  author  = "Koh, Pang Wei and Nguyen, Thao and Tang, Y S and Mussmann, Stephen
             and Pierson, E and Kim, Been and Liang, Percy",
  journal = "ICML",
  volume  = "abs/2007.04612",
  month   =  jul,
  year    =  2020
}

@ARTICLE{Damen2016-gw,
  title    = "Prediction models for cardiovascular disease risk in the general
              population: systematic review",
  author   = "Damen, Johanna A A G and Hooft, Lotty and Schuit, Ewoud and
              Debray, Thomas P A and Collins, Gary S and Tzoulaki, Ioanna and
              Lassale, Camille M and Siontis, George C M and Chiocchia, Virginia
              and Roberts, Corran and Schlüssel, Michael Maia and Gerry, Stephen
              and Black, James A and Heus, Pauline and van der Schouw, Yvonne T
              and Peelen, Linda M and Moons, Karel G M",
  journal  = "BMJ",
  volume   =  353,
  pages    = "i2416",
  month    =  may,
  year     =  2016,
  language = "en"
}

@ARTICLE{Wynants2020-hl,
  title     = "Prediction models for diagnosis and prognosis of covid-19:
               systematic review and critical appraisal",
  author    = "Wynants, Laure and Van Calster, Ben and Collins, Gary S and
               Riley, Richard D and Heinze, Georg and Schuit, Ewoud and Bonten,
               Marc M J and Dahly, Darren L and Damen, Johanna A A and Debray,
               Thomas P A and de Jong, Valentijn M T and De Vos, Maarten and
               Dhiman, Paul and Haller, Maria C and Harhay, Michael O and
               Henckaerts, Liesbet and Heus, Pauline and Kammer, Michael and
               Kreuzberger, Nina and Lohmann, Anna and Luijken, Kim and Ma, Jie
               and Martin, Glen P and McLernon, David J and Andaur Navarro,
               Constanza L and Reitsma, Johannes B and Sergeant, Jamie C and
               Shi, Chunhu and Skoetz, Nicole and Smits, Luc J M and Snell, Kym
               I E and Sperrin, Matthew and Spijker, René and Steyerberg, Ewout
               W and Takada, Toshihiko and Tzoulaki, Ioanna and van Kuijk,
               Sander M J and van Bussel, Bas and van der Horst, Iwan C C and
               van Royen, Florien S and Verbakel, Jan Y and Wallisch, Christine
               and Wilkinson, Jack and Wolff, Robert and Hooft, Lotty and Moons,
               Karel G M and van Smeden, Maarten",
  journal   = "BMJ",
  publisher = "bmj.com",
  volume    =  369,
  pages     = "m1328",
  month     =  apr,
  year      =  2020,
  language  = "en"
}

@ARTICLE{van-Os2022-vv,
  title     = "Developing clinical prediction models using primary care
               electronic health record data: The impact of data preparation
               choices on model performance",
  author    = "van Os, Hendrikus J A and Kanning, Jos P and Wermer, Marieke J H
               and Chavannes, Niels H and Numans, Mattijs E and Ruigrok, Ynte M
               and van Zwet, Erik W and Putter, Hein and Steyerberg, Ewout W and
               Groenwold, Rolf H H",
  journal   = "Front. Epidemiol.",
  publisher = "Frontiers",
  volume    =  2,
  pages     =  871630,
  month     =  jun,
  year      =  2022,
  language  = "en"
}

@ARTICLE{Guevara2024-ul,
  title     = "Large language models to identify social determinants of health
               in electronic health records",
  author    = "Guevara, Marco and Chen, Shan and Thomas, Spencer and Chaunzwa,
               Tafadzwa L and Franco, Idalid and Kann, Benjamin H and Moningi,
               Shalini and Qian, Jack M and Goldstein, Madeleine and Harper,
               Susan and Aerts, Hugo J W L and Catalano, Paul J and Savova,
               Guergana K and Mak, Raymond H and Bitterman, Danielle S",
  journal   = "npj Digital Medicine",
  publisher = "Nature Publishing Group",
  volume    =  7,
  number    =  1,
  pages     = "1--14",
  month     =  jan,
  year      =  2024,
  language  = "en"
}

@ARTICLE{Yang2022-ww,
  title     = "A large language model for electronic health records",
  author    = "Yang, Xi and Chen, Aokun and PourNejatian, Nima and Shin, Hoo
               Chang and Smith, Kaleb E and Parisien, Christopher and Compas,
               Colin and Martin, Cheryl and Costa, Anthony B and Flores, Mona G
               and Zhang, Ying and Magoc, Tanja and Harle, Christopher A and
               Lipori, Gloria and Mitchell, Duane A and Hogan, William R and
               Shenkman, Elizabeth A and Bian, Jiang and Wu, Yonghui",
  journal   = "NPJ Digit. Med.",
  publisher = "Springer Science and Business Media LLC",
  volume    =  5,
  number    =  1,
  pages     =  194,
  month     =  dec,
  year      =  2022,
  language  = "en"
}

@ARTICLE{Agrawal2022-me,
  title     = "Large language models are few-shot clinical information
               extractors",
  author    = "Agrawal, Monica and Hegselmann, Stefan and Lang, Hunter and Kim,
               Yoon and Sontag, David",
  journal   = "Empirical Methods in Natural Language Processing",
  publisher = "Association for Computational Linguistics",
  address   = "Abu Dhabi, United Arab Emirates",
  pages     = "1998--2022",
  month     =  dec,
  year      =  2022
}

@INPROCEEDINGS{Sivaraman2025-it,
  title     = "Tempo: Helping data scientists and domain experts collaboratively
               specify predictive modeling tasks",
  author    = "Sivaraman, Venkatesh and Vaishampayan, Anika and Li, Xiaotong and
               Buck, Brian R and Ma, Ziyong and Boyce, Richard D and Perer, Adam",
  booktitle = "Proceedings of the 2025 CHI Conference on Human Factors in
               Computing Systems",
  publisher = "ACM",
  address   = "New York, NY, USA",
  pages     = "1--18",
  month     =  apr,
  year      =  2025
}

@ARTICLE{Laupacis1997-cj,
  title     = "Clinical prediction rules. A review and suggested modifications
               of methodological standards",
  author    = "Laupacis, A and Sekar, N and Stiell, I G",
  journal   = "JAMA",
  publisher = "American Medical Association (AMA)",
  volume    =  277,
  number    =  6,
  pages     = "488--494",
  month     =  feb,
  year      =  1997,
  language  = "en"
}

@ARTICLE{Wells2000-sc,
  title    = "Derivation of a simple clinical model to categorize patients
              probability of pulmonary embolism: increasing the models utility
              with the {SimpliRED} {D}-dimer",
  author   = "Wells, P S and Anderson, D R and Rodger, M and Ginsberg, J S and
              Kearon, C and Gent, M and Turpie, A G and Bormanis, J and Weitz, J
              and Chamberlain, M and Bowie, D and Barnes, D and Hirsh, J",
  journal  = "Thromb. Haemost.",
  volume   =  83,
  number   =  3,
  pages    = "416--420",
  month    =  mar,
  year     =  2000,
  language = "en"
}

@INCOLLECTION{McLendon2025-sa,
  title     = "Deep venous thrombosis risk factors",
  author    = "McLendon, Kevin and Goyal, Amandeep and Attia, Maximos",
  booktitle = "StatPearls",
  publisher = "StatPearls Publishing",
  address   = "Treasure Island (FL)",
  month     =  jan,
  year      =  2025,
  language  = "en"
}

@ARTICLE{Chung2016-lg,
  title     = "{STOP}-Bang questionnaire: A practical approach to screen for
               obstructive sleep apnea",
  author    = "Chung, Frances and Abdullah, Hairil R and Liao, Pu",
  journal   = "Chest",
  publisher = "Elsevier BV",
  volume    =  149,
  number    =  3,
  pages     = "631--638",
  month     =  mar,
  year      =  2016,
  language  = "en"
}

@ARTICLE{Futoma2020-rc,
  title    = "The myth of generalisability in clinical research and machine
              learning in health care",
  author   = "Futoma, Joseph and Simons, Morgan and Panch, Trishan and
              Doshi-Velez, Finale and Celi, Leo Anthony",
  journal  = "Lancet Digit Health",
  volume   =  2,
  number   =  9,
  pages    = "e489--e492",
  month    =  sep,
  year     =  2020,
  language = "en"
}

@ARTICLE{Hurt2025-gc,
  title     = "The use of an artificial intelligence platform {OpenEvidence} to
               augment clinical decision-making for primary care physicians",
  author    = "Hurt, Ryan T and Stephenson, Christopher R and Gilman, Elizabeth
               A and Aakre, Christopher A and Croghan, Ivana T and Mundi,
               Manpreet S and Ghosh, Karthik and Edakkanambeth Varayil,
               Jithinraj",
  journal   = "J. Prim. Care Community Health",
  publisher = "SAGE Publications",
  volume    =  16,
  pages     =  21501319251332215,
  month     =  jan,
  year      =  2025,
  language  = "en"
}

@ARTICLE{Mattei2020-ea,
  title     = "The story of the development and adoption of the Glasgow Coma
               Scale: Part {I}, the early years",
  author    = "Mattei, Tobias A and Teasdale, Graham M",
  journal   = "World Neurosurg.",
  publisher = "Elsevier BV",
  volume    =  134,
  pages     = "311--322",
  month     =  feb,
  year      =  2020,
  language  = "en"
}

@ARTICLE{Teasdale1974-wx,
  title     = "Assessment of coma and impaired consciousness. A practical scale",
  author    = "Teasdale, G and Jennett, B",
  journal   = "Lancet",
  publisher = "Elsevier",
  volume    =  2,
  number    =  7872,
  pages     = "81--84",
  month     =  jul,
  year      =  1974,
  language  = "en"
}

@INCOLLECTION{Jain2025-tf,
  title     = "Glasgow Coma Scale",
  author    = "Jain, Shobhit and Margetis, Konstantinos and Iverson, Lindsay M",
  booktitle = "StatPearls",
  publisher = "StatPearls Publishing",
  address   = "Treasure Island (FL)",
  month     =  jan,
  year      =  2025,
  language  = "en"
}

@ARTICLE{Jiang2023-sp,
  title    = "Health system-scale language models are all-purpose prediction
              engines",
  author   = "Jiang, Lavender Yao and Liu, Xujin Chris and Nejatian, Nima Pour
              and Nasir-Moin, Mustafa and Wang, Duo and Abidin, Anas and Eaton,
              Kevin and Riina, Howard Antony and Laufer, Ilya and Punjabi,
              Paawan and Miceli, Madeline and Kim, Nora C and Orillac, Cordelia
              and Schnurman, Zane and Livia, Christopher and Weiss, Hannah and
              Kurland, David and Neifert, Sean and Dastagirzada, Yosef and
              Kondziolka, Douglas and Cheung, Alexander T M and Yang, Grace and
              Cao, Ming and Flores, Mona and Costa, Anthony B and
              Aphinyanaphongs, Yindalon and Cho, Kyunghyun and Oermann, Eric
              Karl",
  journal  = "Nature",
  month    =  jun,
  year     =  2023,
  language = "en"
}

@ARTICLE{Seinen2022-zv,
  title     = "Use of unstructured text in prognostic clinical prediction
               models: a systematic review",
  author    = "Seinen, Tom M and Fridgeirsson, Egill A and Ioannou, Solomon and
               Jeannetot, Daniel and John, Luis H and Kors, Jan A and Markus,
               Aniek F and Pera, Victor and Rekkas, Alexandros and Williams,
               Ross D and Yang, Cynthia and van Mulligen, Erik M and Rijnbeek,
               Peter R",
  journal   = "J. Am. Med. Inform. Assoc.",
  publisher = "Oxford University Press (OUP)",
  volume    =  29,
  number    =  7,
  pages     = "1292--1302",
  month     =  jun,
  year      =  2022,
  language  = "en"
}

@ARTICLE{Seinen2025-dc,
  title     = "Using structured codes and free-text notes to measure information
               complementarity in electronic health records: Feasibility and
               validation study",
  author    = "Seinen, Tom M and Kors, Jan A and van Mulligen, Erik M and
               Rijnbeek, Peter R",
  journal   = "J. Med. Internet Res.",
  publisher = "JMIR Publications Inc.",
  volume    =  27,
  number    =  1,
  pages     = "e66910",
  month     =  feb,
  year      =  2025,
  language  = "en"
}

@ARTICLE{Feng2023-ls,
  title     = "Characterization of risk prediction models for acute kidney
               injury: A Systematic Review and Meta-analysis",
  author    = "Feng, Yunlin and Wang, Amanda Y and Jun, Min and Pu, Lei and
               Weisbord, Steven D and Bellomo, Rinaldo and Hong, Daqing and
               Gallagher, Martin",
  journal   = "JAMA Netw. Open",
  publisher = "American Medical Association",
  volume    =  6,
  number    =  5,
  pages     = "e2313359",
  month     =  may,
  year      =  2023,
  language  = "en"
}
}

\newpage

\appendix

\setcounter{table}{0}
\setcounter{figure}{0}
\renewcommand{\thetable}{A\arabic{table}}
\renewcommand{\thefigure}{A\arabic{figure}}
\renewcommand{\theHfigure}{AppendixFigure\arabic{figure}}
\renewcommand{\theHtable}{AppendixTable\arabic{table}}
{\section{Extended Data}}
\FloatBarrier




\begin{table}
\centering
\caption{\textbf{Complete list of concepts for TBI prediction across \method iterations.}
In round 1, we additionally consider running \method in an outcome-agnostic way, where the LLM does not know which outcome it is predicting and must rely on the prediction model to identify useful features.
Nevertheless, the model successfully identifies useful features, even recovering a few of the original PECARN concepts.
}
\label{tab:tbi_concepts}
\footnotesize
\begin{tabular}{>{\raggedright\arraybackslash}p{2cm}p{1cm}p{13cm}}
\toprule
\textbf{Round} & \textbf{Coef.} & \textbf{Concepts} \\
\midrule
\textbf{Round 1}
 & 2.70 & Does the note mention the patient experiencing a loss of consciousness? \\
 & 1.74 & Does the note mention a brain bleed? \\
 & 1.58 & Does the note mention the patient having a neurological event? \\
 & 1.23 & Does the note mention a Glasgow Coma Scale score? \\
 & 1.10 & Does the note mention the patient being seizure-free? \\
\midrule
\textbf{Round 2}  & 3.46 & Does the note mention the patient experiencing loss of consciousness? \\
 & 1.50 & Does the note mention the patient having a normal Glasgow Coma Scale score and being at least 2 years old? \\
 & 1.04 & Does the note mention the patient experiencing convulsions? \\
 & 0.82 & Does the note mention the patient having altered mental status? \\
 & -0.65 & Does the note mention the patient having intact cranial nerves? \\
\midrule
\textbf{Round 3}  & 4.34 & Does the note mention the patient experiencing loss of consciousness? \\
 & 1.41 & Does the note mention the patient having a history of mild TBI? \\
 & 1.21 & Does the note mention the patient having an occipital hematoma? \\
 & 1.07 & Does the note mention the patient having vision changes? \\
 & -0.53 & Does the note mention the patient having memory intact? \\
 \midrule
  \textbf{Round 4} & 1.58 & Does the note mention the patient experiencing unconsciousness? \\
 & 0.73 & Does the note mention the patient having altered mental status? \\
 & 0.25 & Does the note mention the patient experiencing a headache? \\
 & 0.08 & Does the note mention the patient experiencing head trauma? \\
 & -0.22 & Does the note mention the patient having normal gait? \\
\bottomrule
\end{tabular}
\end{table}

\begin{table}
    \centering
    \caption{\textbf{Comparing the sensitivity and  specificity of TBI models from HACHI and comparator methods.}
    To make the models comparable, the threshold is chosen so that the sensitivity is as close to 0.90 as possible.
    95\% CI is shown in parentheses.
    }
    \begin{tabular}{p{1.5cm} p{2cm} p{2cm} p{2cm} p{2cm} p{2cm} p{2cm}}
    \toprule
        Model & Baseline (PECARN)& OpenEvidence+ LLMs & HACHI (Round 1) & HACHI (Round 2) & HACHI (Round 3) & HACHI (Round 4)\\
        \hline
        Sensitivity & 0.942 (0.942, 1.000) & 0.913 (0.900, 1.000) & 0.906 (0.900, 0.993) & 0.913 (0.900, 0.938) & 0.906 (0.900, 0.950) & 0.906 (0.900, 0.938)\\
        \hline
        Specificity & 0.556 (0.000, 0.613) & 0.642 (0.000, 0.884) & 0.481 (0.154, 0.906) & 0.615 (0.510, 0.875) & 0.749 (0.332, 0.891) & 0.807 (0.545, 0.883)
        \\
        \bottomrule
    \end{tabular}
    \label{tab:tbi_specificity}
\end{table}

\begin{table}
\centering
\caption{\textbf{Complete list of concepts for AKI prediction across \method iterations.}}
\begin{tabular}{>{\raggedright\arraybackslash}p{0.7cm}p{0.7cm}p{15cm}}
\toprule
\textbf{Round} & \textbf{Coef.} & \textbf{Concepts} \\
\midrule
 \textbf{1} & 1.31 & Does the note mention this patient having leukocytosis? \\
 & 0.87 & Does the note mention this patient having abdominal distention? \\
 & 0.74 & Does the note mention this patient having cardiac dysfunction? \\
 & 0.71 & Does the note mention this patient having tachycardia? \\
 & 0.65 & Does the note mention this patient experiencing swelling? \\
 & 0.57 & Does the note mention this patient having a systemic infection? \\
 & 0.53 & Does the note mention this patient having renal impairment? \\
 & 0.47 & Does the note mention this patient having diabetes mellitus? \\
 & -0.36 & Does the note mention this patient having no kidney disease? \\
 & -1.01 & Does the note mention this patient having low hematocrit? \\
\midrule
\textbf{2}   & 1.28 & Does the note mention the patient having chronic kidney disease? \\
 & 1.03 & Does the note mention the patient undergoing an exploratory laparotomy? \\
 & 0.98 & Does the note mention the patient having sepsis? \\
 & 0.89 & Does the note mention the patient having fluid retention? \\
 & 0.47 & Does the note mention the patient having a respiratory condition? \\
 & 0.43 & Does the note mention the patient requiring a blood transfusion during surgery? \\
 & 0.40 & Does the note mention the surgery being high-risk? \\
 & 0.37 & Does the note mention the patient having heart disease? \\
 & 0.31 & Does the note mention the patient having a malignancy? \\
 & -1.01 & Does the note mention the patient having minimally invasive surgery? \\
\midrule
\textbf{3}   & 1.05 & Does the note mention the patient having chronic kidney disease, e.g., a history of CKD or elevated creatinine levels? \\
 & 0.91 & Does the note mention the surgery being major or high-risk, e.g., major abdominal surgery or anticipated blood loss $>$ 500mL? \\
 & 0.88 & Does the note mention the patient having tachycardia, e.g., heart rate $>$ 100 bpm or palpitations? \\
 & 0.72 & Does the note mention the patient having an active systemic infection, e.g., sepsis, bacteremia, or requiring therapeutic antibiotics? \\
 & 0.64 & Does the note mention the patient having heart failure, e.g., reduced ejection fraction or history of heart failure exacerbations? \\
 & 0.51 & Does the note mention the patient having a history of sleep apnea, e.g., witnessed apnea or CPAP use? \\
 & 0.50 & Does the note mention the surgery being urgent or emergent, e.g., requiring immediate attention or intervention? \\
 & 0.39 & Does the note mention the patient having obesity, e.g., high BMI or overweight status? \\
 & 0.31 & Does the note mention the patient having hypertension, e.g., high blood pressure or antihypertensive medication use? \\
 & -0.74 & Does the note mention the surgery being minimally invasive, e.g., laparoscopic or robotic-assisted procedures? \\
\bottomrule
\end{tabular}
\label{tab:aki_concepts}
\footnotesize
\end{table}

\pagebreak
\section{Example prompts for TBI in Round 1}

Here we show examples of prompts used for TBI in Round 1; see all prompts on \href{http://github.com/jjfenglab/HACHI}{Github}.
\footnotesize

\subsection{KeyphrasePrompt}
\begin{Verbatim}
You are extracting clinical descriptors of a pediatric patient from emergency room medical notes that may influence their risk of being diagnosed with traumatic brain injury (TBI).  Focus entirely on factors explicitly mentioned in the notes. Do not infer or add details unless explicitly stated.

Extract factors that may:
- Increase TBI risk, be specific and detailed (e.g., substance use, prior head injuries)
- Decrease TBI risk, be specific and detailed (e.g., protective equipment use, medication compliance)
- Indicate TBI severity or complications, be specific and detailed (e.g., neurological symptoms, cognitive changes)

For each factor, provide:
1. The primary descriptor (exact or paraphrased from the note)
2. Synonyms or alternative phrasings
3. Broader generalizations that capture similar concepts

Requirements:
- Each descriptor should be $\leq$3 words
- The descriptor should state distinguishing features.
- List specific terms, synonyms, and increasingly general concepts. However, do not state overly general concepts like "protective factor", "risk factor", "low risk", or "high risk" because we are using these terms to brainstorm/learn specific factors that clinicians can use to predict TBI risk.
- Separate items with commas


Note(s):
"{note}"

Do not infer or fabricate details. Base all responses strictly on the medical note text.

Output Format:
{
    "reasoning": "Describe any reasoning here",
    "keyphrases": [
        "Primary descriptor, Synonym1, Synonym2, Broader category",
        "Primary descriptor, Alternative phrasing, More general concept",
        ...
    ]
}

Examples:
{
    "reasoning": "Describe any reasoning here",
    "keyphrases": [
        "Crowded residence, Overcrowded living, Constrained environment, Environmental risk",
        "Cervical strain, Neck pain, Muscle spasm",
        "Physical assault, Blunt trauma, Injury infliction",
        ...
    ]
}
\end{Verbatim}

\subsection{ConceptInitializationProposalPrompt}
\begin{Verbatim}
We are fitting a concept bottleneck model to predict traumatic brain injury (TBI) for pediatric patients admitted to emergency room. The goal is to come up with a few meta-concepts that clinicians can use to easily risk-stratify patients for whether they are at high risk of being diagnosed with TBI. As training data, we are using clinical notes to retrospectively extract patient characteristics.

The top words most associated with TBI were estimated using a simple logistic regression model with tf-idf features. Based on the top words, come up with {max_meta_concepts} meta-concepts. A meta-concept has to be defined in terms of a yes/no question, e.g. "Does the note mention this patient having a fall?". Pick meta-concepts that are as specific as possible, without being too broad.

Suggestions for generating candidate meta-concepts: Do not propose meta-concepts that are simply a union of two different concepts (e.g. "Does the note mention this patient having an infection or being elderly?" is not allowed), questions with answers that are almost always a yes (e.g. the answer to "Does the note mention this patient being hospitalized?" is almost always yes), or questions where the yes/no options are not clearly defined (e.g. "Does the note mention this patient experiencing difficulty?" is not clearly defined because difficulty may mean financial difficulty, physical difficulties, etc).

Words with top coefficients:

{top_features_df}

Propose at least {max_meta_concepts} candidates in the following JSON format:
{
  "reasoning": "Step-by-step explanation of how you systematically worked through the model concepts (from most to least predictive).”,
  "concepts": [
    {
      "concept": "<QUESTION 1>",
      "words": ["<SYNONYM/ANTONYM 1>", "<SYNONYM/ANTONYM 2>", "<SYNONYM/ANTONYM 3>"]
    },
    {
      "concept": "<QUESTION 2>",
      "words": ["<SYNONYM/ANTONYM 1>", "<SYNONYM/ANTONYM 2>", "<SYNONYM/ANTONYM 3>"]
    },
    ...
    {
      "concept": "<QUESTION {max_meta_concepts}>",
      "words": ["<SYNONYM/ANTONYM 1>", "<SYNONYM/ANTONYM 2>", "<SYNONYM/ANTONYM 3>"]
    }
  ]
}

Example answer:
{
  "reasoning": "Step-by-step reasoning through the model concepts",
  "concepts": [
    {
      "concept": "Does the note mention the patient experiencing head trauma?",
      "words": ["head injury", "skull trauma", "cranial damage"]
    },
    {
      "concept": “Does the note mention loss of consciousness?”
      "words": ["LOC", "fainting, unresponsive"]
    },
    ...
    ]
}
\end{Verbatim}

\subsection{ConceptReplacementProposalPrompt}
\begin{Verbatim}
The goal is to come up with a concept bottleneck model (CBM) that extracts {num_concepts} meta-concepts from clinical notes from the electronic health record of pediatric patients admitted to the emergency room to predict risk of having a traumatic brain injury (TBI). A meta-concept is a binary feature extractor defined by a yes/no question. We have {num_concepts_fixed} meta-concepts so far:
{meta_concepts}

To come up with the {num_concepts}th meta-concept, I have done the following: I first fit a CBM on the {num_concepts_fixed} existing meta-concepts. Then to figure out how to improve this {num_concepts_fixed}-concept CBM, I first extracted a list of concepts that are present in each note, and then fit a linear regression model on the extracted concepts to predict the residuals of the {num_concepts_fixed}-concept CBM. These are the top extracted concepts in the resulting residual model, in descending order of importance:

{top_features_df}

To interpret this residual model, a general rule of thumb is that an extracted concept with a large positive coefficient means that a high risk of TBI is positively associated with the concept being mentioned in the note, and an extracted concept with a negative coefficient means that a higher risk of TBI is negatively associated with the concept being mentioned in the note. 

Given the residual model, create cohesive candidates for the {num_concepts}th meta-concept. Be systematic and consider all the listed concepts in the residual model. Start from the most to the least predictive concept. For each concept, check if it matches an existing meta-concept or create a new candidate meta-concept. Work down the list, iterating through each concept. Clearly state each candidate meta-concept as a yes/no question.

Suggestions for generating candidate meta-concepts: Do not propose meta-concepts that are simply a union of two different concepts (e.g. "Does the note mention this patient having an infection or being elderly?" is not allowed), questions with answers that are almost always a yes (e.g. the answer to "Does the note mention this patient being hospitalized?" is almost always yes), or questions where the yes/no options are not clearly defined (e.g. "Does the note mention this patient experiencing difficulty?" is not clearly defined because difficulty may mean financial difficulty, physical difficulties, etc). Do not propose meta-concepts where you would expect over 95% agreement or disagreement with the {num_concepts_fixed} existing meta-concepts (e.g. "Does the note mention the patient having homelessness?" overlaps too much with "Does the note mention the patient having housing insecurity?").

Finally, summarize all the generated candidates for the {num_concepts}-th meta-concept in a JSON. Merge any candidate meta-concepts that are essentially the same (where you would expect over 95% agreement) or essentially opposites (you would expect over 95% disagreement). In the JSON, include a list of comma-separated list of phrases that mean the same or opposite of each candidate meta-concept. Propose at least ten candidates. The final JSON should have the following format:

Propose at least six candidates in the following JSON format:
{
  "reasoning": "Step-by-step explanation of how you systematically worked through the model concepts (from most to least predictive) and combined them with the feedback suggestions.”,
  "concepts": [
    {
      "concept": "<QUESTION 1>",
      "words": ["<SYNONYM/ANTONYM 1>", "<SYNONYM/ANTONYM 2>", "<SYNONYM/ANTONYM 3>"]
    },
    {
      "concept": "<QUESTION 2>",
      "words": ["<SYNONYM/ANTONYM 1>", "<SYNONYM/ANTONYM 2>", "<SYNONYM/ANTONYM 3>"]
    },
    ...
    {
      "concept": "<QUESTION 6>",
      "words": ["<SYNONYM/ANTONYM 1>", "<SYNONYM/ANTONYM 2>", "<SYNONYM/ANTONYM 3>"]
    }
  ]
}

Example answer:
{
  "reasoning": "Step-by-step reasoning through the residual model concepts and feedback",
  "concepts": [
    {
      "concept": "Does the note mention the patient experiencing head trauma?",
      "words": ["head injury", "skull trauma", "cranial damage"]
    },
    {
      "concept": “Does the note loss of consciousness?”
      "words": ["LOC", "fainting, unresponsive"]
    },
    ...
    ]
}
\end{Verbatim}

\subsection{ConceptAnnotationPrompt}
\begin{Verbatim}
You will be given clinical notes. I will give you a series of questions. Your task is answer each question with a probability from 0 to 1. Summarize the response with a JSON that includes your answer to all of the questions. Questions:
{prompt_questions}

clinical notes:
{sentence}

Example answer: {
    "0": 0,
    "1": 1,
    "2": 0.5
}
Answer all the questions and do not answer with anything else besides valid JSON. Do not add comments to the JSON.
\end{Verbatim}

\end{document}